\documentclass[times,twocolumn,final]{elsarticle}
\usepackage{style_arxiv}
\usepackage{framed,multirow}
\usepackage{breqn}
\usepackage{siunitx}
\usepackage{amssymb}
\usepackage{latexsym}

\usepackage{url}
\usepackage{xcolor}
\usepackage{graphicx} 
\usepackage{pifont}
\usepackage{makecell}
\usepackage{enumitem}
\usepackage[titletoc]{appendix}
\usepackage{placeins}
\usepackage{framed,multirow}
\usepackage{subcaption}

\makeatletter
\newcommand*{\rom}[1]{\expandafter\@slowromancap\romannumeral #1@}
\makeatother

\usepackage{amsmath} 
\usepackage{xcolor}

\usepackage[colorlinks=true,urlcolor=green]{hyperref}

\makeatletter
\newcommand{\bianca}{\renewcommand\NAT@open{[}\renewcommand\NAT@close{]}}
\makeatother

\definecolor{newcolor}{rgb}{.8,.349,.1}

\begin{document}

\verso{Luca Sestini \textit{et~al.}}

\begin{frontmatter}

\title{SAF-IS: a Spatial Annotation Free Framework for Instance Segmentation of Surgical Tools}

\author[1,2]{Luca \snm{Sestini}\corref{cor1}}
\cortext[cor1]{Corresponding author: }
  \ead{sestini@unistra.fr}
\author[1]{Benoit \snm{Rosa}}
\author[2]{Elena \snm{De Momi}}
\author[2]{Giancarlo \snm{Ferrigno}}
\author[1,3]{Nicolas \snm{Padoy}}

\address[1]{ICube, University of Strasbourg, CNRS, France}
\address[2]{Department of Electronics, Information and Bioengineering, Politecnico di Milano, Milano, Italy}
\address[3]{IHU Strasbourg, Strasbourg, France}

\begin{abstract}

Instance segmentation of surgical instruments is a long-standing research problem, crucial for the development of many applications for computer-assisted surgery. This problem is commonly tackled via fully-supervised training of deep learning models, requiring expensive pixel-level annotations to train.\\
In this work, we develop a framework for instance segmentation not relying on spatial annotations for training. Instead, our solution only requires binary tool masks, obtainable using recent unsupervised approaches, and binary tool presence  labels, freely obtainable in robot-assisted surgery. Based on the binary mask information, our solution learns to extract individual tool instances from single frames, and to encode each instance into a compact vector representation, capturing its semantic features. Such representations guide the automatic selection of a tiny number of instances (8 only in our experiments), displayed to a human operator for tool-type labelling. The gathered information is finally used to match each training instance with a binary tool presence label, providing an effective supervision signal to train a tool instance classifier.\\
We validate our framework on the EndoVis 2017 and 2018 segmentation datasets. We provide results using binary masks obtained either by manual annotation or as predictions of an unsupervised binary segmentation model. The latter solution yields an instance segmentation approach completely free from spatial annotations, outperforming several state-of-the-art fully-supervised segmentation approaches.
\end{abstract}

\end{frontmatter}

\section{Introduction}
\label{sec:introduction}

\begin{figure*}[t]
    \centerline{\includegraphics[width=\linewidth]{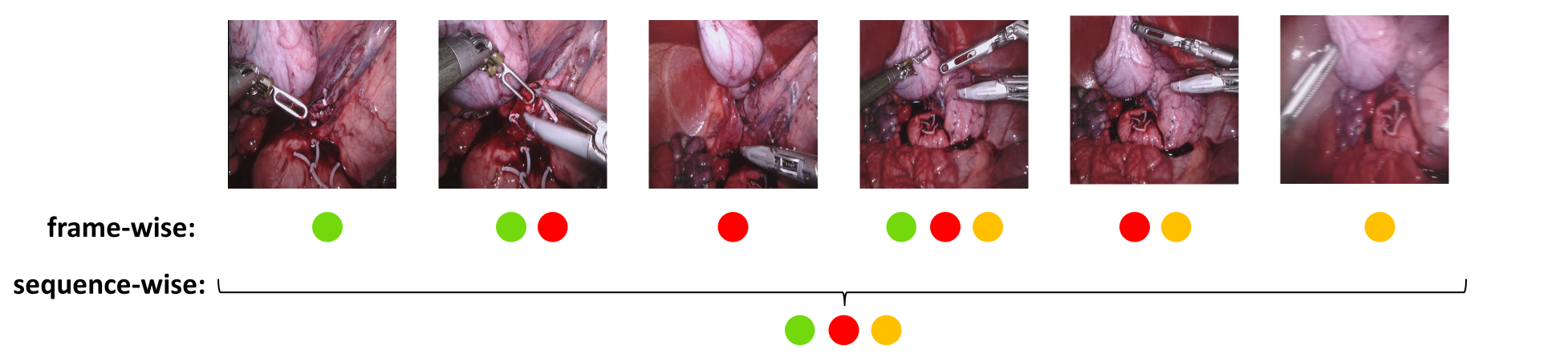}}
    \caption{Examples of \textit{frame-wise} and \textit{sequence-wise} binary tool presence labels for a robot-assisted surgery sequence from the EndoVis 2017 dataset (each color represents a tool type). All the tools can be attached to the system at the same time, while being visible only in certain frames.}
    \label{fig:SAF-IS:fr_seq_labs}
\end{figure*}
Endoscopic videos from minimally-invasive procedures offer rich information describing the surgical act. The automatic analysis of such information opens up several opportunities to better understand surgical practice and to improve it \citep{francis2018eaes,lavanchy2021automation,mascagni2022artificial}. Surgical computer vision provides the necessary tools to process raw endoscopic videos, enabling the extraction of dense information for downstream applications. Among the various surgical computer vision tasks, automatic instrument localisation and identification represent an essential component of many downstream applications, like surgical skill assessment \citep{lavanchy2021automation}, augmented reality \citep{tanzi2021real}, 3D scene reconstruction \citep{wang2022neural} and 3D pose estimation \citep{allan20183}. This problem is often formalized by means of either \textit{semantic} or \textit{instance} segmentation. Semantic Segmentation (SeS) aims at directly labelling each image pixel as either belonging to the \textit{background} class or to a certain \textit{tool type} class. Instance Segmentation (IS) aims at localising and identifying individual tool instances, providing, for each instance, a separate mask and a tool-type class labels. Such \textit{tool instantiation} information, \textit{i.e.} the availability of a separate segmentation mask for each tool instance present in the image, is extremely precious for downstream applications like automatic skill assessment, as it enables individual tool tracking over time. State-of-the-art approaches commonly tackle tool segmentation via fully-supervised training of deep learning models \citep{shvets2018automatic,jin2019incorporating,kong2021accurate,kurmann2021mask}. Such approaches require the availability of pixel-level semantic and instance labels, extremely expensive to collect via manual annotation at a large scale. This confines the training of such models to small annotated datasets, limiting their generalization ability.\\
Recently, alternatives to standard fully-supervised approaches have been proposed for the task of binary tool segmentation, a type of SeS featuring only two classes, \textit{tool} and \textit{background} \citep{sahu2020endo,sestini2023fun,pakhomov2020towards,sestini2021kinematic,da2019self}. Most of these solutions rely on semi-synthetic dataset generation, for example by combining simulation data and domain translation approaches \citep{sahu2020endo}. While appealing, their application is still potentially limited by the domain gap between synthetic and real data, and by the need for ad-hoc setups to collect synthetic data. A few alternative works have shown the potential of prior instrument knowledge to train deep learning models for binary segmentation, without requiring spatial annotations. This has been done by exploiting prior knowledge on instrument motion and shape \citep{sestini2023fun}, or, in the robotic context, by incorporating 3D tool models and kinematic data \citep{pakhomov2020towards,sestini2021kinematic,da2019self}.\\ Despite these growing efforts to reduce the dependency on manual annotations, research has remained confined to the binary segmentation task. We believe that this is due to the rigid problem formalization imposed by common instance and semantic segmentation approaches: such approaches do not benefit from the potential availability of binary segmentation masks, as they would still require pixel-level labels to train. Furthermore, this problem formalization prevents the incorporation of significantly cheaper sources of semantic information, compared to spatial annotations, like binary tool presence labels. Specifically, we define as \textit{frame-wise} these binary tool presence labels describing which tool types are \textit{effectively} visible in each frame; we define as \textit{sequence-wise} these labels indicating which tool types are \textit{potentially} visible in each frame (see Figure \ref{fig:SAF-IS:fr_seq_labs} for examples of a sequence). While \textit{frame-wise} labels are usually obtained via manual annotation - although much cheaper than spatial annotation - \textit{sequence-wise} labels can be automatically obtained from different sources. In robot-assisted surgery, for example, robotic systems can often record which tools are attached \citep{kurmann2021mask}. This information only indicates that a certain tool could be visible at some point while it is attached, but does not guarantee its visibility in any specific frame (therefore a \textit{sequence-wise} visibility). As a generalization, surgical phase and step annotations could provide similar information, when a mapping between phases/steps and tools can be approximately defined, for example by knowing which tools are commonly used in each phase/step \citep{padoy2012statistical}. While the use of \textit{frame-wise} labels has been explored by weakly-supervised tool detection methods \citep{vardazaryan2018weakly,nwoye2019weakly}, no segmentation solutions have yet included them in their training pipelines. Furthermore, to the best of our knowledge, no approaches have yet explored the use of automatically obtainable \textit{sequence-wise} labels.\\
\begin{figure*}[t]
    \centerline{\includegraphics[width=\textwidth]{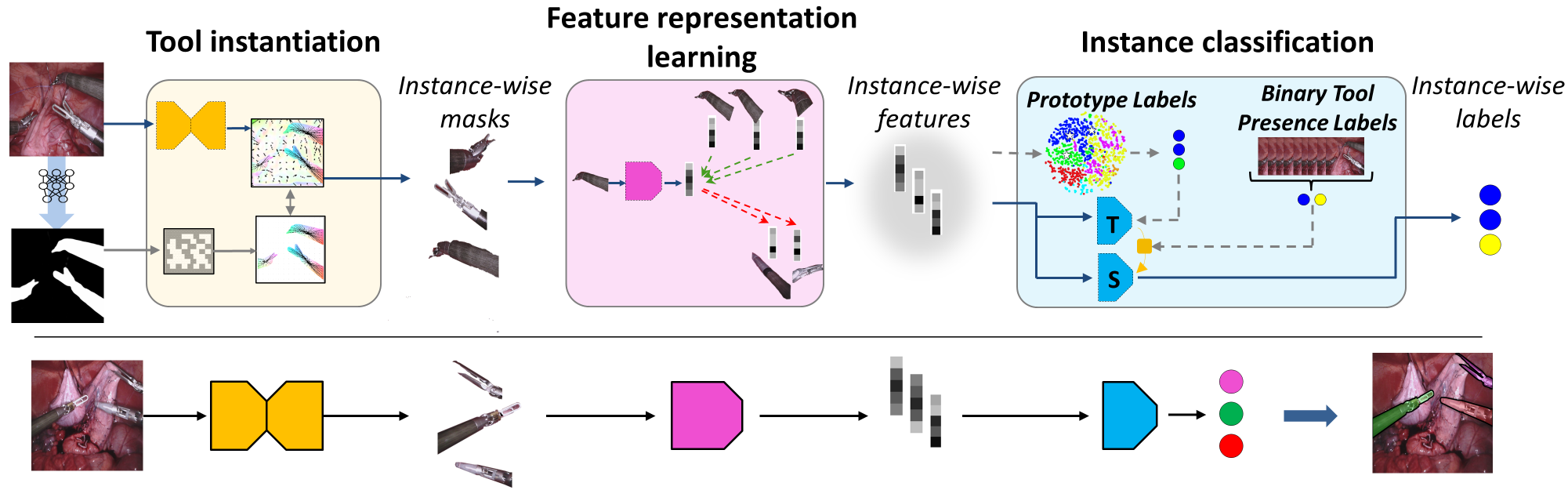}}
    \caption{Overview of the proposed Spatial Annotation Free framework for Instance Segmentation (SAF-IS). Top: training architecture highlighting the three core steps. \textit{Tool instantiation} is learnt from binary masks, potentially obtained using recent unsupervised segmentation methods. \textit{Feature representation learning} is performed using a contrastive learning strategy, powered by local temporal tracking. This step allows us to extract a feature representation of each tool instance in the training set. \textit{Instance classification} is performed by incorporating a minimal amount of human-provided information (\textit{prototype labels}, as few as 8 in our experiments) and cheaply obtainable binary tool presence labels. Bottom: SAF-IS inference architecture.}
    \label{fig:SAF-IS:fullmethod}
\end{figure*}
\indent In this work we propose a framework for instance segmentation model training, which embraces the recent progress on unsupervised binary segmentation and the availability of cheap binary tool presence labels, either \textit{frame-wise} or \textit{sequence-wise}. 
Compared to pixel-level annotations, binary tool presence labels are not spatially localized. 
Weakly-supervised tool detection approaches \citep{vardazaryan2018weakly,nwoye2019weakly} exploit the class-activation maps provided by a classifier trained on the \textit{frame-wise} binary tool presence labels, in order to localize the tools in the image space. However, such localization is commonly limited to discriminant parts of the tools, like the tip, thus not suitable for segmentation. In addition, such approaches cannot handle \textit{sequence-wise} labels, as these labels do not provide a ground truth signal for the training of the frame-wise classifier. To tackle these challenges our solution first learns to localize individual tool instances and to encode each of them in a compact feature representation. These instance-wise representations are then used to select a small number of tool instances (\textit{prototype instances}, 8 only in our experiments), which are presented to a human operator for tool-type labelling. The gathered information is finally used to match each instance to a semantic label from the corresponding set of binary tool presence labels, providing an effective supervision signal for the training of an instance classifier.\\
To this aim, we make the following contributions:
\begin{itemize}
    \item we develop an unsupervised approach for tool instantiation (Figure \ref{fig:SAF-IS:fullmethod}, \textit{Tool instantiation}). This step allows training a model to extract a separate binary segmentation mask for each tool instance present in a frame. With no availability of pixel-level labels, we fabricate a pseudo-supervision signal from the connected component instantiation of the binary masks, and refine it using simple assumptions on instrument positioning in the image space. This signal is used to train the instantiation model, directly predicting the position of each instance centroid in the image space in the form of a 2D displacement field;
    
    \item we develop a self-supervised approach for feature representation learning (Figure \ref{fig:SAF-IS:fullmethod}, \textit{Feature representation learning}): this step allows training a model to encode each tool instance in a compact representation, capturing its semantic features. With no availability of pixel-level semantic labels, we learn such representations by relying on intrinsic temporal information from video sequences. Specifically, we design a contrastive learning approach based on local instance tracking to draw positive and negative samples. This step allows obtaining powerful instance-wise feature representations, providing the necessary information to solve the final classification training step;
    
    \item we develop an approach to learn instance classification from the binary tool presence labels (Figure \ref{fig:SAF-IS:fullmethod}, \textit{Instance  classification}). The feature representations of all the training instances, learnt at the previous step, are used to guide the automatic selection of a tiny number of prototype instances, displayed to a human operator for tool-type labelling. The gathered information is propagated to the whole training set, allowing us to label each training instance with a pseudo tool-type label (\textit{prototype labels}).  This information is combined with the available binary tool presence labels (either \textit{frame-wise} or \textit{sequence-wise}) using a teacher-student approach. This step allows matching each training instance to a semantic label from the corresponding set of binary tool presence labels, providing an effective supervision signal for the training of the student instance classifier;

    \item at inference time the trained architecture can perform instance segmentation on single frames, by extracting individual tool instances, encoding each of them in a compact feature representation, and separately classifying them (Figure \ref{fig:SAF-IS:fullmethod}, bottom).\\
\end{itemize}

\section{Related Work}
\label{rw}
Surgical instrument segmentation is a long-standing research problem. Before the Deep Learning (DL) break-through, the problem was tackled by totally relying on prior knowledge about surgical tools, like color distribution \citep{wei1997automatic}, shape \citep{bouget2015detecting} or orientation in the field-of-view \citep{voros2006automatic}. \\
Following the DL irruption in the field, a great research effort has been dedicated to designing powerful fully-supervised architectures, boosting segmentation accuracy. Such solutions are presented below in Section \ref{rw:fully-supervised}, focusing in particular on instance segmentation approaches. Although fully-supervised methods have achieved unprecedented segmentation results on benchmark datasets, their scalability is restricted by the need for manual annotations, which confines their training to small annotated datasets, limiting their generalization ability. To address this challenge, various approaches have been suggested, which we present in Section \ref{rw:non-fully-supervised}. 

\subsection{Fully Supervised Solutions}
\label{rw:fully-supervised}
Following the DL breakthrough in the field of surgical computer vision, research works have mostly addressed the problem of surgical tool segmentation using fully-supervised DL approaches. In particular, encoder-decoder architectures based on Convolutional Neural Networks (CNNs) have been widely adopted, in concurrency with a semantic segmentation formulation of the problem. \citep{garcia2017toolnet,shvets2018automatic,pakhomov2019deep,hasan2019u} propose different variations of the U-Net architecture \citep{ronneberger2015u}, exploring different loss functions, residual connections, dilated convolutions and ad-hoc augmentation pipelines. Multi-task learning has also been adopted, coupling the segmentation task with image-based localisation of tool landmarks \citep{laina2017concurrent} and task-oriented saliency maps prediction \citep{islam2021st}. While the segmentation task can be solved for single frames, temporal information has been proven to boost performance, especially in the case of partially occluded tools \citep{jin2019incorporating}.\\
\indent Recently, instance segmentation approaches have started gaining traction. Several of the proposed approaches are based on the popular Mask-RCNN architecture \citep{he2017mask}. \citep{kong2021accurate} directly train a Mask-RCNN architecture for the task of surgical instrument instance segmentation. ISI-Net \citep{gonzalez2020isinet} adds a temporal-consistency module for improved segmentation results. \citep{kurmann2021mask} propose a \textit{mask-then-classify} approach, adopting an anchor-free approach for instrument instantiation, based on direct localisation of instruments centroids. Differently from the above-listed methods, \citep{zhao2022trasetr} simultaneously tackle the problems of instance segmentation and tracking using a transformer architecture based on the popular TrackFormer and DETR models \citep{carion2020end,meinhardt2022trackformer}.\\
In this work we also adopt an instance segmentation problem formalization, showing its benefits beyond fully-supervised training.

\subsection{Non Fully-Supervised Solutions}
\label{rw:non-fully-supervised}
Motivated by the need to reduce the burden of manual annotation, several solutions have tackled the segmentation problem by including unlabelled data in the training process, exploiting small sets of labelled data, weak annotations or prior knowledge. Such solutions, mostly focusing on the binary segmentation problem, are presented below. \\
\textbf{Semi-Supervised solutions:} this family of approaches incorporates unlabelled data in the training process, while still requiring access to a set of manually annotated data. Different solutions to combine unlabelled and labelled data have been explored. \citep{ross2018exploiting} pre-train a CNN on unlabelled data, by means of a pretext task carried out using a cycle-GAN architecture, and then fine-tune the model on annotated data. A similar pipeline can be followed by replacing the pre-text task with self-supervised representation learning on the unlabelled data, as experimentally proven by \citep{ramesh2022dissecting}. \citep{zhao2020learning} tackle the problem of sparsely annotated data, propagating low hertz annotations to intermediate unlabelled frames using optical-flow. \citep{kalia2021co} incorporate unlabelled data from different domains in the training process to improve generalization to these domains. This is achieved by mapping annotated frames from the labelled set to the unlabelled domain using a cycle-GAN architecture, allowing for better generalization.\\
\textbf{Weakly-Supervised solutions}:
weakly-supervised training is a learning paradigm trying to solve a task using annotations cheaper to obtain compared to the ones required by the fully-supervised paradigm. Such annotations could be a simplification of the ideal ground truth annotations, like scribbles in place of masks for the segmentation task, or annotations providing indirect/incomplete supervision for the targeted task (e.g. binary tool presence labels for tool localisation tasks).
For segmentation, the application of weakly-supervised training remains confined to the binary task. \citep{lee2019weakly} propose a framework to integrate scribble-like annotations, speeding up the annotation process. \citep{yang2022weakly} automatically obtain a pseudo-supervision signal by attaching an electromagnetic sensor to the surgical instruments. While cutting the cost of annotations, the approach is inherently limited by regulatory constraints, which limit the extent of validation of this study.\\
Weak annotations, in the form of \textit{frame-wise} binary tool presence labels, have mostly been used to tackle the problem of bounding-box localisation. \citep{vardazaryan2018weakly} train a multi-label classifier to predict tool presence from single frames; the designed architecture features an extended spatial pooling layer yielding class-specific feature maps, used during inference to localise the tools. Similarly \citep{nwoye2019weakly} use Wildcat Pooling \citep{durand2017wildcat} to obtain localisation maps, adding a convolution-LSTM module for improved temporal consistency. Differently from these two approaches, \citep{xue2022new} use binary tool presence labels, in combination with green-screen recorded images of surgical instruments, to obtain a pseudo-supervision signal consisting of noisy and redundant bounding boxes. A bounding-box regressor is then trained on the noisy supervision signal, and its predictions for a certain tool are averaged together according to their confidence score.\\
The use of \textit{frame-wise} binary tool presence labels has remained limited to the bounding-box detection task, as the standard approach involving using class-activation maps limits the localisation to discriminative parts of the tools, missing out significant parts of the instruments like the shafts. Furthermore, research works on weakly-supervised learning have mostly focused on \textit{frame-wise} labels, which still require a certain annotation effort. This has led to overlooking the opportunity given by \textit{sequence-wise} binary tool presence labels, particularly valuable, for example, in robot-assisted surgery.\\
\textbf{Prior knowledge based solutions}: as proven by early works on tool segmentation, general assumptions on color distribution of endoscopic frames, instrument position and prior shape knowledge, can be a sufficient source of information to localise surgical instruments. \citep{liu2020unsupervised}, for example, generate segmentation pseudo-labels using handcrafted cues, such as color distribution; binary segmentation results are then refined exploiting feature correlation between adjacent video frames. \citep{sestini2023fun} propose FUN-SIS, an approach exploiting general assumptions on instrument motion and \textit{shape-priors} to train a binary segmentation model, achieving results comparable to the ones of fully-supervised solutions. \\ 
\indent In this work we combine the use of prior knowledge and binary tool presence labels to learn instance segmentation of surgical instruments. Prior knowledge on instrument positioning in the field-of-view is exploited to instantiate binary segmentation masks. Weak information, in the form of binary tool presence labels, both \textit{frame-wise} and \textit{sequence-wise}, is then incorporated to achieve accurate instance classification.


\section{Methodology}
\label{Sect:SAF-IS:method}
The proposed SAF-IS framework for Spatial Annotation Free Instance Segmentation explicitly separates the task into three core components: tool instantiation, feature representation and instance classification. Differently from standard semantic/instance segmentation approaches, SAF-IS does not require spatial annotation of the training data. Instead, it relies on the availability of binary segmentation masks, which can be cheaply obtained using emerging unsupervised approaches, and binary tool presence labels.\\
The full framework is presented in Figure \ref{fig:SAF-IS:fullmethod} and detailed below.

\subsection{Tool Instantiation}
\label{Sect:SAF-IS:method:inst}

Instrument instantiation is here defined as the problem of predicting, from an endoscopic image $I$, the set of binary masks $\{M^{Inst}_i\}$, with $i$ in $[1,N_{Inst}]$, each one corresponding to an individual instrument visible in the image. When the ground truth instantiation is known, the problem is often formulated as bounding-box prediction \citep{kong2021accurate,gonzalez2020isinet}. However, the effectiveness of this approach has been questioned in \citep{kurmann2021mask}, which proposed an alternative solution based on direct regression of instance centroids' position. We here adopt a similar formulation, showing its benefits with respect to bounding-box prediction beyond fully-supervised learning.\\
The instantiation problem is here formalized as learning the mapping between the image $I \in R^{W\times H\times3}$ and the displacement field $D \in R^{W\times H\times2}$, uniquely assigning each tool pixel to an instance. Given a pixel $\mathbf{p} = [p_x,p_y]$, $D|_{\mathbf{p}}$ is equal to the vector $\mathbf{v} = [c^i_x-p_x,c^i_y-p_y]$ if $\mathbf{p}$ belongs to a certain instance $i$, having its centroid in $[c^i_x,c^i_y]$, or to the null vector $[0,0]$, if $\mathbf{p}$ belongs to the background. Given a training set with known ground truth instantiation $D$, such mapping can be learnt by an instantiation model, implemented as a neural network, by using a fully-supervised training formulation, as in \citep{kurmann2021mask}. This can be achieved by optimizing the loss $L^{FS}_I$, implemented as the pixel-wise distance between the ground truth displacement field $D$ and the instantiation model prediction $\tilde{D}$:

\begin{equation}
    L^{FS}_I = |D - \tilde{D}|.
\end{equation}

At inference time, given a new image $I$ and the corresponding predicted displacement field $\tilde{D}$, the set of instance masks $\{M^{Inst}_i\}$ can be easily extracted by identifying the instance centroids, as the pixels where the displacement field converges, and assigning each tool pixel to the centroid pointed by the corresponding displacement vector.

\begin{figure}[t]
    \centerline{\includegraphics[width=3.5in]{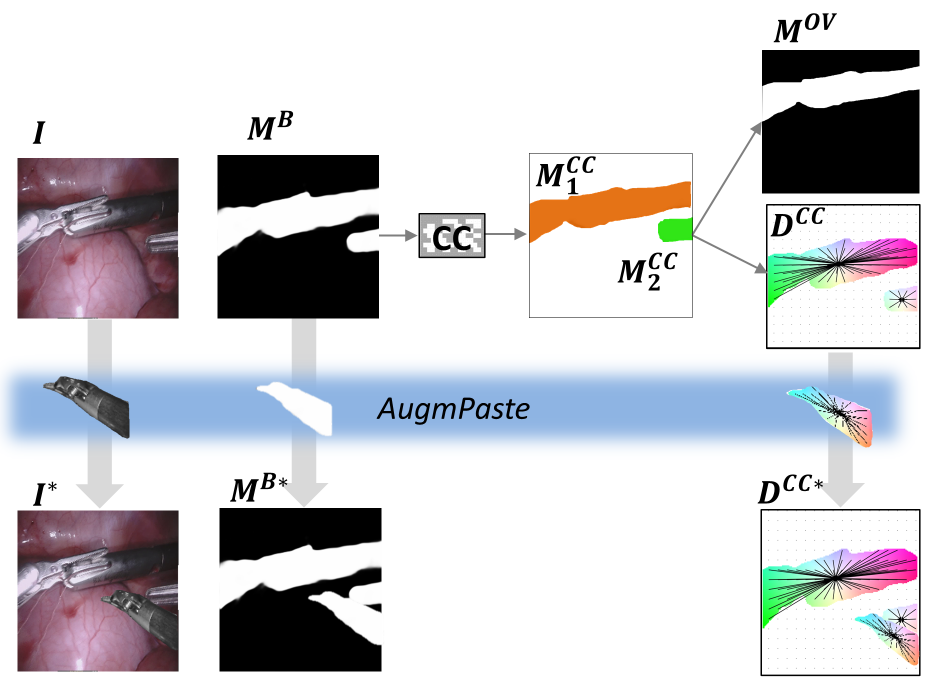}}
    \caption{Overview of the proposed strategy to generate a pseudo-supervision signal to learn instrument instantiation. Given an image $I$, its binary mask $M^B$ is instantiated using a Connected Component (CC) algorithm, yielding the set of tool masks $\{M^{CC}_i\}$, with $i$ in $[1,N_{CC}]$. From them, the displacement field $D^{CC}$ and the overlap mask $M^{OV}$ can be automatically obtained. A random tool instance is then selected from the training set, and pasted on $I$, $M^B$, $D^{CC}$, producing their augmented versions $I^{\mathbf{*}}$, $M^{B\mathbf{*}}$, $D^{CC\mathbf{*}}$.}
    \label{fig:SAF-IS:inst_training}
\end{figure}

\begin{figure*}[ht]
    \centerline{\includegraphics[width=\linewidth]{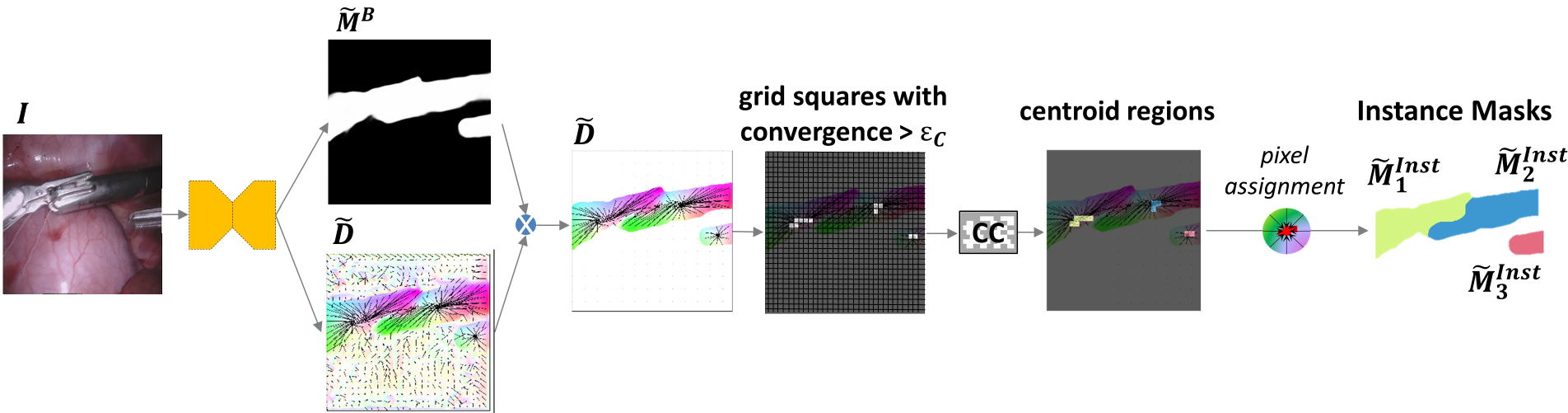}}
    \caption{Overview of the proposed instantiation strategy. Given an image $I$ the trained instantiation model predicts the masked displacement field $\tilde{D}$. A square grid is then overlapped to $\tilde{D}$, and the squares with high convergence (per-pixel average $> \varepsilon_C$) are then extracted, and separated by Connected Component (CC) labelling, yielding a set of $\tilde{N}_{Inst}$ centroid regions. Each tool pixel is then assigned to the corresponding centroid, yielding the set of instance masks $\{\tilde{M}^{Inst}_i\}$, with $i$ in $[1,\tilde{N}_{Inst}]$.}
    \label{fig:SAF-IS:inst_inf}
\end{figure*}

\textbf{Training:} In our case, only the binary mask $M^B$ is known. Without a ground truth instantiation, we rely on the assumption that surgeons tend to avoid overlapping surgical instruments in the field-of-view, in order to reduce the chances of mutual tool occlusions and unwanted tool interactions.\\
Given an image $I$ and the corresponding binary mask $M^B$, if tools do not overlap, the instance masks can be obtained by separating the Connected Components (CC) of $M^B$ through standard computer vision methods like the Spaghetti algorithm \citep{bolelli2019spaghetti}. The displacement field $D^{CC}$, approximating the ground truth $D$, can then be directly obtained from the set of $N_{CC}$ tool masks $\{M^{CC}_i\}$, with $i$ in $[1,N_{CC}]$, by subtracting each tool pixel position from the centroid $[c^i_x,c^i_y]$ of the corresponding mask $M^{CC}_i$.
While effective in the case of non-overlapping tools, CC labelling systematically fails when tools overlap. In order to mitigate this problem we artificially modify the supervision signal obtained from CC instantiation, as follows:
\begin{itemize}
    \item \textbf{potential overlapping tools identification}: in minimally invasive surgery surgeons adopt the principle of \textit{triangulation} to increase their ability to visualize and access anatomy \citep{russo2012triangulation}. As a result, surgical tools commonly enter the camera's field-of-view from the sides. Therefore, given the set of CC masks $\{M^{CC}_i\}$, $M^{CC}_i$ is considered a potential overlapping instance if it covers the full horizontal length of the frame (see Figure \ref{fig:SAF-IS:inst_training} for an example). All pixels corresponding to potential overlapping instances are collected in the binary overlap mask $M^{OV}$, and discarded from loss computation as described later in this Section;
    \item \textbf{instance pasting augmentation (\textit{AugmPaste})}: given an image $I$, its binary mask $M^B$ and its CC displacement field $D^{CC}$, a random tool instance is selected from a different training sample and pasted on them, yielding the augmented image $I^{\mathbf{*}}$, the augmented binary mask $M^{B\mathbf{*}}$ and the augmented displacement field $D^{CC\mathbf{*}}$ (Figure \ref{fig:SAF-IS:inst_training}). This augmentation step allows us to artificially simulate the presence of overlapping instances, making up for the discarded instances at the previous step.
\end{itemize}

Given the image $I^*$, in addition to the displacement field $\tilde{D}$, we let the instantiation model predict the binary segmentation mask $\tilde{M}^B$, which we multiply by $\tilde{D}$ to ensure that the displacement vector for pixels belonging to the background is a null vector $[0,0]$. For simplicity, we keep the notation $\tilde{D}$ to refer to the result of such product.\\
Given the image $I^{\mathbf{*}}$, the corresponding network predictions $\tilde{D}$ and $\tilde{M}^{B}$, the binary mask $M^{B\mathbf{*}}$, the displacement field $D^{CC\mathbf{*}}$ and the overlap mask $M^{OV}$, the instantiation model is trained by optimizing the loss $L_I$:
\begin{equation}
    L_I = |D^{CC\mathbf{*}} - \tilde{D}|(1-M^{OV}) + L_{CE}(M^{B\mathbf{*}},\tilde{M}^{B}),
\end{equation}
where $L_{CE}$ is a standard pixel-wise cross-entropy loss. 

\begin{figure*}[t]
    \centerline{\includegraphics[width=0.9\linewidth]{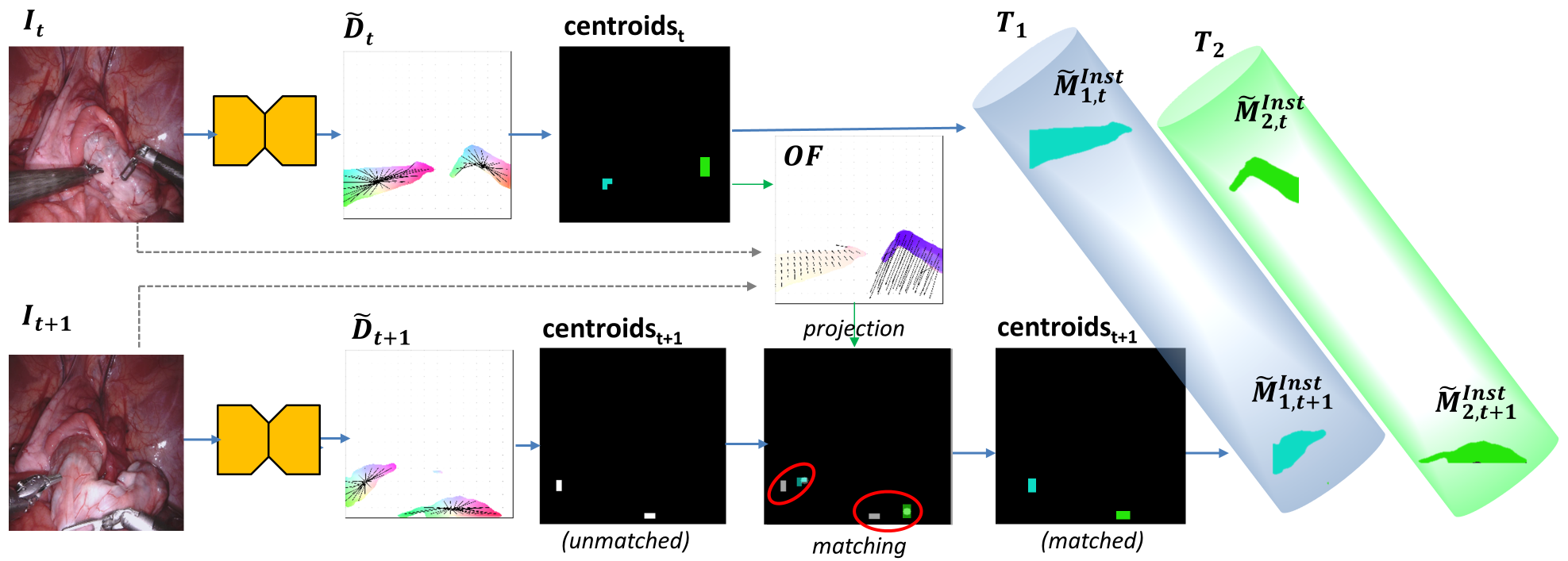}}
    \caption{Overview of the tracking strategy used to generate positive samples for contrastive learning. Given two consecutive frames, the centroids at time $t$, obtained from the displacement field $D_t$ are mapped to the $I_{t+1}$ space using optical flow $OF$, computed between the two images $I_t$ and $I_{t+1}$. The projected centroids are then matched to the ones obtained from the displacement field $D_{t+1}$. This allows building the set of tubes $\{T_i\}$, with $i$ in $[1,\tilde{N}_{Inst}]$. Tubes are progressively grown by repeating this process for consecutive frames.}
    \label{fig:SAF-IS:tracking}
\end{figure*}

\indent \textbf{Inference:} given an image $I$ and the trained instantiation model, the predicted displacement field $\tilde{D}$ must me mapped to the set of instance masks $\{\tilde{M}^{Inst}_i\}$, with $i$ in $[1,\tilde{N}_{Inst}]$, and $\tilde{N}_{Inst}$ being the number of predicted instances in a frame.  While for the ground truth displacement field $D$ each tool pixel vector points exactly to the corresponding centroid pixel, this is not guaranteed for the predicted $\tilde{D}$. Therefore we define as \textit{centroids} the regions of $\tilde{D}$ with a high rate of displacement vectors convergence. Practically, we overlap a square grid to $\tilde{D}$ and compute, for each square, the per-pixel average number of vectors pointing inside it. If such number is above a predefined threshold $\varepsilon_C$, the square is considered a centroid square. Connected squares are grouped together, to yield the set of centroid regions $\{c_i\}$, with $i$ in $[1,\tilde{N}_{Inst}]$. 
The instance masks can then be extracted by assigning each tool pixel $\mathbf{p}$ to the centroid $c_i$ closest to the point identified by $\mathbf{p} + \tilde{D}|_{\mathbf{p}}$. This yields the set of predicted instance masks $\{\tilde{M}^{Inst}_i\}$, with $i$ in $[1,\tilde{N}_{Inst}]$ (Figure \ref{fig:SAF-IS:inst_inf}). In our framework the predicted instance masks are subsequently used to learn instance-wise feature representations, as now discussed.

\begin{figure*}[t]
    \centerline{\includegraphics[width=6.in]{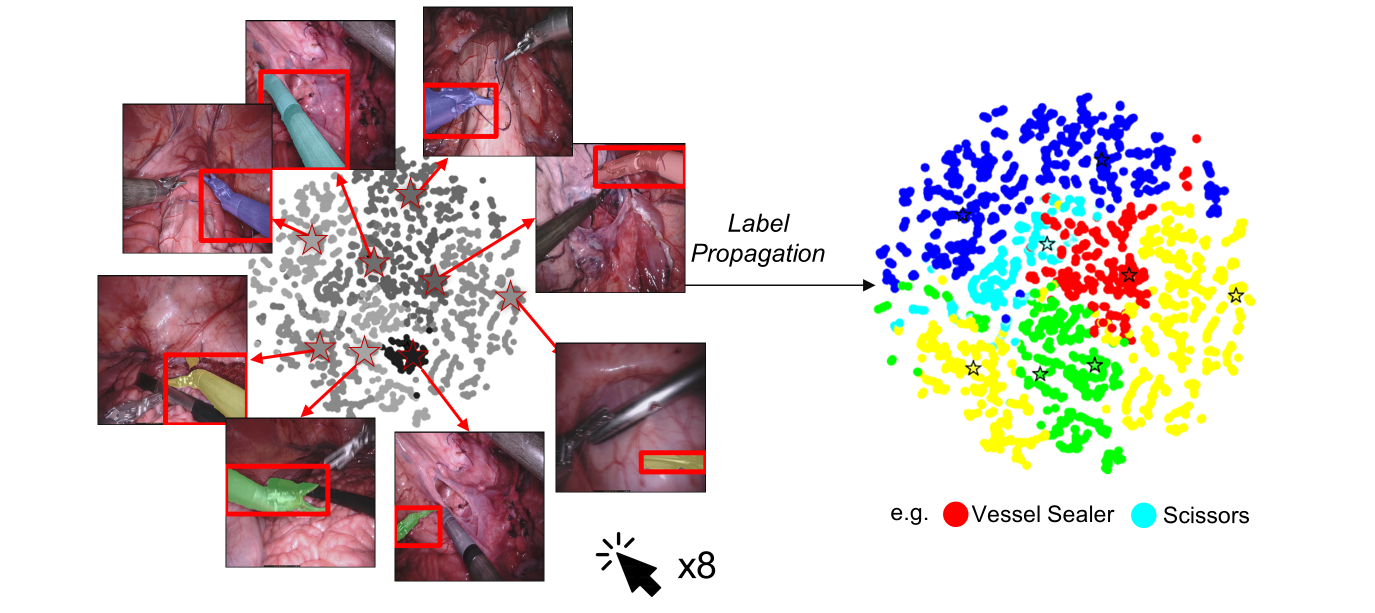}}
    \caption{Left: visualization of learnt feature representations of the EndoVis 2017 \citep{allan20192017} training set instances, clustered ($N_{km}$ equal to 8) and projected in the 2D space using t-SNE algorithm \citep{van2008visualizing}. Each instance point is colored with a different shade of grey to represent the cluster id. Prototype instance features are marked with $\star$, and the corresponding masks are overlaid on the frame and highlighted by a bounding-box, to facilitate their labelling by a human operator. The color of the mask overlays represents the ground truth tool type that the user would assign. In practice, this annotation step can be carried out in \textit{8 mouse clicks} only. Right: prototype instance labels propagation to the training set. Each instance-wise feature projection is colored accordingly to its prototype label, assigned via propagation from the prototype instance.}
    \label{fig:SAF-IS:interface}
\end{figure*}

\subsection{Feature Representation Learning}
\label{Sect:SAF-IS:method:featlearn}

In the absence of pixel-level semantic labels, we rely on self-supervision to learn robust and meaningful feature representations of each tool instance, tailored for the instance segmentation task. The problem of self-supervised representation learning has been often addressed by means of contrastive learning in literature \citep{jaiswal2020survey}. While general contrastive learning approaches usually learn global frame-level feature representations, we find this formulation to be ill-posed for the instrument segmentation problem, as it lacks the spatial granularity necessary to discriminate between different instances.
Therefore we design an instance-level contrastive learning approach, exploiting the unsupervised instantiation described above and intrinsic temporal information from video sequences. \\
Given an image $I$ and the set of instance masks $\{\tilde{M}^{Inst}_i\}$ predicted by the instantiation model, we want to map each instance to a feature vector $F_i$, capturing its semantic content. We obtain feature vectors using a feature extractor model implemented using a standard ResNet-50 architecture. Specifically, for each instance, we pass $I$ through the model and multiply the intermediate feature maps by ${\tilde{M}^{Inst}_i}$, resized to match their dimensions, to obtain the corresponding instance-wise feature vector $F_i$. Then, given a feature representation $F_i$, intrinsic temporal information from the video sequence is used to draw positive and negative examples for contrastive loss computation. Specifically:
\begin{itemize}
    \item positive examples $\{F_i^+\}$ are sampled from the instance tube $T_{i}$, built from the frame-by-frame tracking of the instance $i$. Such tracking is described in Figure \ref{fig:SAF-IS:tracking}. Given the consecutive images $I_t$ and $I_{t+1}$, and their corresponding sets of instrument instances, tracking is solved by projecting the centroids of $I_t$ into $I_{t+1}$ space using the optical flow $OF$, computed between $I_t$ and $I_{t+1}$. Each $I_t$ centroid is then matched to the closest $I_{t+1}$ centroid. Optical flow projection allows us to robustly handle tool movements between consecutive frames, reducing the chances of wrong matching;  
    \item negative examples $\{F_i^-\}$ can be sampled either from different tubes belonging to the same frame or from tubes far apart in time. 
\end{itemize}

\noindent The feature extractor network is then trained by optimizing the loss $L_F$ between $\{F_i^+\}$ and $\{F_i^-\}$:
\begin{equation}
    L_F = L_{SCL}(\{F_i^+\},\{F_i^-\}),
\end{equation}
where $L_{SCL}$ is the Supervised Contrastive Loss formulation proposed in \citep{khosla2020supervised}, with each instance tube treated as a separate class. The learnt feature representations are exploited in the next step for classifier training.

\begin{figure*}[t]
    \centerline{\includegraphics[width=\textwidth]{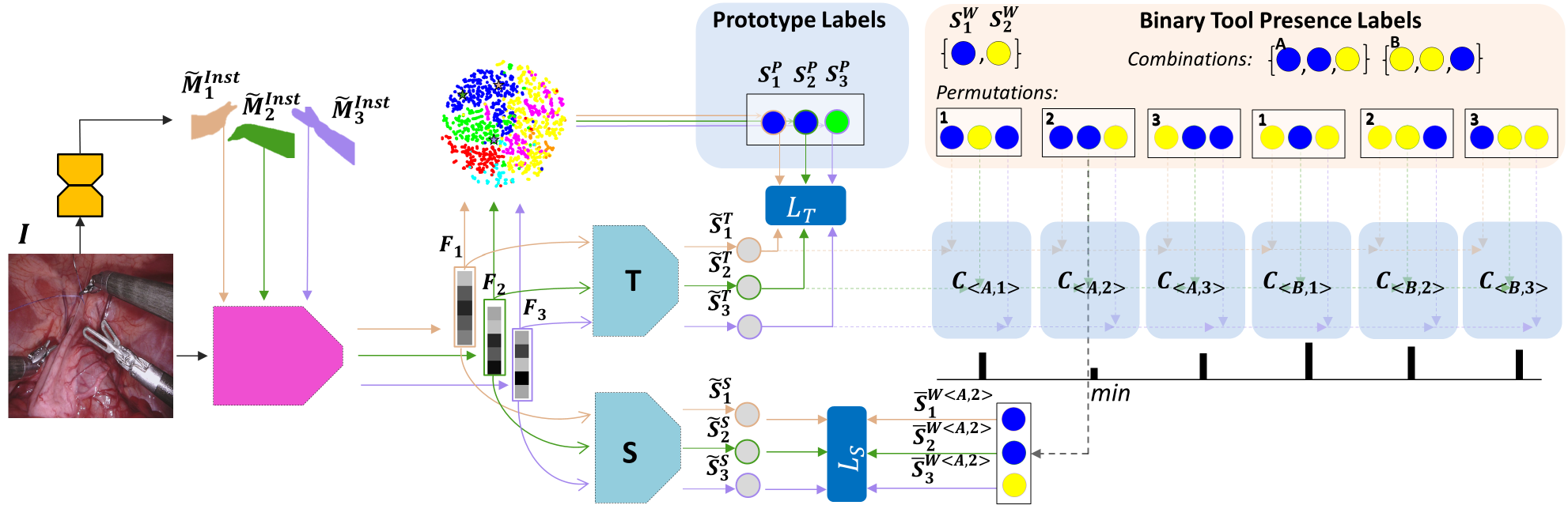}}
    \caption{Overview of the proposed weakly-supervised instance classification module. Given an image $I$, the corresponding set of instance-wise features $\{F_i\}$, with $i$ in $[1,\tilde{N}_{Inst}]$, is obtained from the instance masks $\tilde{M}^{Inst}_i$. Each feature is mapped to the corresponding prototype label $S^P_i$, which, as shown in this case, does not necessarily correspond to the ground truth label. Each feature is also independently passed through the Teacher (\textbf{T}) and Student networks (\textbf{S}), yielding the predicted probabilities $\mathbf{\tilde{P}^T_i}$, $\mathbf{\tilde{P}^S_i}$, and the corresponding predicted labels $\tilde{S}^T_i$, $\tilde{S}^S_i$ (for the sake of readability only the latter are shown in the picture). \textbf{T} is trained optimizing the loss $L_T$ computed using the prototype labels $\{S^P_i\}$. Simultaneously, \textbf{T} predictions are used to compute the assignment costs $C_{\langle i_C,i_P \rangle}$ for each $i_P$ permutation of each $i_C$ combination of the weak labels $\{S^W_i\}$, with $i$ in $[1,N_W]$. The ordered set $[\bar{S}^W_i]$, with $i$ in $[1,\tilde{N}_{Inst}]$, corresponding to the minimum assignment cost, is used to compute the loss $L_S$ for Student network optimization.
    \label{fig:SAF-IS:classifier_training}}
\end{figure*}

\subsection{Instance classification}
\label{Sect:SAF-IS:method:class}
Given the set of available tool type classes $\{S_i\}$, with $i$ in $[1,N_{cls}]$, a classifier model must now be trained to learn the mapping between instance features and class labels from that set. In the absence of pixel-level semantic labels, we rely on binary tool presence labels to solve this task, cheaper to collect via manual annotation, or even automatically obtainable (e.g. \textit{sequence-wise} labels from robotic systems). As binary tool presence labels are not spatially localised, the matching between training tool instances and binary tool presence labels must be defined. The class-activation approach, commonly adopted for weakly-supervised object detection, requires frame-wise ground-truth annotations about tool presence, which makes it inapplicable to \textit{sequence-wise} labels. We therefore propose a more flexible solution, applicable to both \textit{frame-wise} and \textit{sequence-wise} labels. Our solution is designed to solve the matching problem by injecting a minimal amount of human knowledge, specifically collected to maximize its information content while minimizing the annotation effort. Specifically, we automatically select a tiny number of highly representative instances (\textit{protoype} instances) and ask a human operator to label them. The gathered information is then used to match binary tool presence labels and instances, providing an effective supervision signal for classifier training. The two steps are now detailed.\\
\indent \textbf{Prototype labelling:} given the complete set of learnt features for all the instances in the training set, unsupervised clustering is applied. In our experiments we make use of the standard K-Means++ clustering algorithm \citep{arthur2006k}, with the number of clusters $N_{km}$ regarded as an hyper-parameter. The $N_{km}$ instances corresponding to the clusters' centroids are defined as \textit{prototype instances}. A human operator would now be required to assign a label $S^P$ from the set $\{S_i\}$ to each prototype instance. In order to propagate the prototype instance labels to the rest of the training instances, we require all instances belonging to the same cluster to share the same semantic label $S^P$. Figure \ref{fig:SAF-IS:interface} provides a visualization the of prototype instance labelling process, and of the result of prototype label propagation.
In principle, a number of clusters $N_{km}$ equal to $N_{cls}$, the total number of tool type classes available, is sufficient to correctly label the whole training dataset, and potentially to directly deploy the instance segmentation model: given an unseen image $I$ and a predicted tool instance mask $\tilde{M}^{Inst}_i$ from that image, inference would then be performed by extracting the corresponding feature vector $F_i$ and associating it to the prototype label $S^P$ of the cluster closest to $F_i$ in the feature space. However, in practice, as the feature learning step is imperfect, the prototype labels can be noisy, as experimentally shown in Section \ref{Sect:SAF-IS:res}. Nonetheless, we show that the information provided by prototype labels can be used to match binary tool presence labels and instances, providing an effective supervision signal for classifier training.

\indent \textbf{Binary tool presence labels incorporation}: let us consider the set of binary tool presence labels $\{S^W_i\}$ with $i$ in $[1,N_{W}]$, subset of the set of tool-type labels $\{S_i\}$, associated to a certain frame. As discussed in Section \ref{sec:introduction}, this information can be defined as \textit{frame-wise}, if the labels indicate which tool types are \textit{effectively} visible in the frame, or \textit{sequence-wise}, if they indicate which tool types are visible at some point in the sequence the frame belongs to, but not necessarily in such frame. Binary tool presence information, either \textit{frame-wise} or \textit{sequence-wise}, does not provide tool localisation information, and is therefore defined as \textit{weak} with respect to the segmentation task. While cheaply obtainable, such weak labels are often overlooked by segmentation approaches, as they pose several challenges:
\begin{itemize}
    \item differently from pixel-level labels, binary tool presence labels are not directly matched to a specific instance, making them hard to digest for standard segmentation architectures, designed to learn from pixel-level annotations;
    \item depending on the system/annotation protocol used to collect the information, the presence of multiple instances of the same tool type may not be recorded. In the Cholec80 dataset \citep{twinanda2016endonet}, for example, \textit{frame-wise} binary tool presence labels do not keep track of multiple tool instances;
    \item \textit{sequence-wise} labels commonly do not reflect which tool types are effectively visible in each frame. In the case of robotic surgery, for example, tools are attached beforehand to the robotic system, potentially remaining unused for relatively long periods of time. Similarly for surgical phases, certain tools, like the ones used for coagulation, may be linked to every phase of a procedure, while being visible only for small amounts of time.
\end{itemize}
In order to make effective use of such information, each tool instance in a frame must be matched to a weak label from the set $\{S^W_i\}$ associated to that frame. Once the matching is found, a classifier model can be trained on the matched labels. In practice, the binary tool presence labels softly constrain the training of the classifier, providing a reduced set of tool-type labels among which the ground truth one for each instance is to be found.\\
\indent Let us consider an image $I$, the sets of instance masks, features and  prototype labels $\{\tilde{M}^{Inst}_i\},\{F_i\},\{S^P_i\}$, with $i$ in $[1,\tilde{N}_{Inst}]$, and the set of weak labels $\{S^W_i\}$, with $i$ in $[1,N_W]$ associated to $I$. Mining such weak labels requires to find the function $\xi$, matching the set of $\tilde{N}_{Inst}$ features to the set of $N_W$ weak labels. However, in the most general case, such transformation is:
\begin{itemize}
    \item \textit{non injective}, as there could be multiple instances sharing the same tool label $S^W_i$;
    \item \textit{non surjective}, as a certain tool label $S^W_i$ may not be present in a specific frame.
\end{itemize}

\noindent This implies that given the set of $\tilde{N}_{Inst}$ tool instances in a frame, different combinations of $\tilde{N}_{Inst}$ elements of the $N_W$ weak labels are plausible. To simplify the problem, and avoid degenerate solutions, we assume that if the number of instances $\tilde{N}_{Inst}$ in a frame is equal or smaller than the number of weak labels for that frame, every instance is assigned to a different label. Specifically, we identify the set of plausible weak labels combinations as follows:
\begin{itemize}
    \item if $\tilde{N}_{Inst} < N_{W}$, all the possible combinations of $\tilde{N}_{Inst}$ elements of the $N_W$ labels are plausible;
    \item if $\tilde{N}_{Inst} == N_W$, we assume that the set of $N_W$ labels is the only plausible combination;
    \item if $\tilde{N}_{Inst} > N_W$, all the possible combinations with repetitions of $\tilde{N}_{Inst}$ elements of the $N_W$ labels are plausible.
\end{itemize}

\noindent Among the set of plausible weak labels combinations, the correct label combination must be identified, and the matching between each instance and each weak label in such combination must be determined. This could be achieved by associating to each $i_P$ permutation of each $i_C$ plausible combination of the weak labels an assignment cost $C_{\langle i_C,i_P \rangle}$. Each couple $\langle i_C,i_P \rangle$ yields an ordered set of weak labels $[S^{W\langle i_P,i_C \rangle}_i]$, with $i$ in $[1,\tilde{N}_{Inst}]$. Among them, the ordered set minimizing the assignment cost could be selected and used for the classifier training.\\
To solve this problem we propose a teacher-student approach (Figure \ref{fig:SAF-IS:classifier_training}), exploiting the knowledge gathered from the prototype labels. Teacher and Student are two identical classifiers that map a feature vector $F_i$ to the vectors $\mathbf{\tilde{P}^T_i}$, $\mathbf{\tilde{P}^S_i}$, respectively. $\mathbf{\tilde{P}^T_i}$, $\mathbf{\tilde{P}^S_i}$ represent the predicted probability of the instance to belong to each of the $N_{cls}$ classes, according to Teacher and Student, respectively. From $\mathbf{\tilde{P}^T_i}$, $\mathbf{\tilde{P}^S_i}$ the class with the highest probability $\tilde{S}^T_i$, $\tilde{S}^S_i$ is regarded as the predicted label. The Teacher network is trained to map each feature $F_i$ to the corresponding prototype label $S^P_i$, by optimizing the instance classification loss $L_{T_i}$:
\begin{equation}
    L_{T_i} = L_{CE}(\mathbf{\tilde{P}^T_i},S^P_i).
\end{equation}

 \noindent For each couple $\langle i_C,i_P \rangle$, its assignment cost $C_{\langle i_C,i_P \rangle}$ can then be computed as the average cross-entropy loss between the predicted probabilities $[\mathbf{\tilde{P}^T_i}]$ and the weak labels $[S^{W\langle i_P,i_C \rangle}_i]$, corresponding to that couple, as follows:
\begin{equation}
    C_{\langle i_C,i_P \rangle} = \frac{1}{\tilde{N}_{Inst}}\sum_{i=1}^{\tilde{N}_{Inst}} L_{CE}(\mathbf{\tilde{P}^T_i},S^{W\langle i_C,i_P\rangle}_i).
\end{equation}

\noindent The ordered set of weak labels $[\bar{S}^{W,i_P,i_C}_i]$, corresponding to the couple $\langle i_C,i_P \rangle$ minimizing the assignment cost, is selected. The Student network is then trained by optimizing the instance classification loss $L_{S_i}$, between the predicted probabilities $[\mathbf{\tilde{P}^S_i}]$ and the matched weak labels $[\bar{S}^{W,i_P,i_C}_i]$:
\begin{equation}
    L_{S_{i}} =  L_{CE}(\mathbf{\tilde{P}^S_i},\bar{S}^{W\langle i_C,i_P\rangle}_i).
\end{equation}

In practice, the Teacher network applies the knowledge gathered from the prototype labels to identify the correct ordered set of weak labels used for Student training. Doing so, the Teacher approximates the function $\xi$, matching each of the $\tilde{N}_{Inst}$ tool instances to a weak label from the set $\{S^W_i\}$. \\
This general framework applies to both \textit{frame-wise} and \textit{sequence-wise} binary tool presence labels. In the case of \textit{frame-wise} labels, $\xi$ becomes surjective, significantly reducing the space of possible solutions and facilitating the matching. 

X
\section{Experimental Set-up}
The proposed framework was validated on the MICCAI 2017 and 2018 EndoVis Robotic Instrument Segmentation Challenge datasets. The two datasets are now introduced (Section \ref{Sect:SAF-IS:exp:data}), together with the specific design choices and training details (Section \ref{Sect:SAF-IS:exp:design}).
\subsection{Datasets}
\label{Sect:SAF-IS:exp:data}
\indent \textbf{EndoVis2017} \citep{allan20192017}: the original challenge dataset consists of 10 video clips, resampled at a frame rate of 1 frame-per-second, of abdominal porcine procedures, performed using the da Vinci robotic system. Each clip contains 300 high-resolution frames (1024 $\times$ 1280). During the challenge 8x225 frames were released for training, while the remaining 8x75 frames and two additional clips were held out by the organizers for testing. A total of 7 tool classes are present in the dataset. We provide results on this dataset according to the same evaluation protocol as \citep{shvets2018automatic}, by performing 4-fold cross-validation on the 8x225 released training data (regrouped in 4 splits). We report the average metric over the 4 splits, for direct comparison with state-of-the-art approaches. \\
\indent \textbf{EndoVis2018} \citep{allan20202018}: the original challenge dataset contains 19 video clips, resampled at a frame rate of 1 frame-per-second, of abdominal porcine procedures, performed using da Vinci robotic system. Each video contains a total of 300 high-resolution frames (1024 $\times$ 1280). During the challenge 15 clips were released for training, while the remaining clips were held out by the organizers for testing. The dataset was originally annotated for anatomy and tool-part segmentation, and did not feature instrument type labels. \citep{gonzalez2020isinet} annotated with pixel-level semantic labels 149 frames for each of the 15 training clips, and split them into a  training set consisting of 11 clips, and a validation set containing the remaining 4 clips. The same 7 tool classes from EndoVis2017 dataset were used. We provide results on this dataset according to the same evaluation protocol as \citep{gonzalez2020isinet}, by training on the 11 training clips, and validating on the remaining 4 clips.\\

As the proposed SAF-IS approach requires binary instrument masks to train, we provide results using both manually annotated binary masks and automatically segmented masks generated using the unsupervised FUN-SIS approach \citep{sestini2023fun}. The mean binary IoU for the FUN-SIS approach on the EndoVis2017 and EndoVis2018 datasets is equal to 83.7\% and 81.3\%, respectively.\\
\indent \textit{Frame-wise} binary tool presence labels were automatically generated for each frame as the unique pixel-level semantic labels present in the corresponding ground truth semantic masks. \textit{Sequence-wise} binary tool presence labels were also automatically generated, by considering each video clip in the datasets as a sequence, and assigning to each clip, as \textit{sequence-wise} labels, the full set of unique semantic labels present in the ground truth semantic masks of all the frames in the clip. For 46.12\% of the frames in the EndoVis2018 dataset the \textit{sequence-wise} labels do not correspond to the \textit{frame-wise} labels (40.72\% for EndoVis2017 dataset), i.e., for a certain frame, its \textit{sequence-wise} labels contain at least a tool type which is not visible in it (but which is present at some point in the clip it belongs to).

\begin{table*}[t]
\centering
\begin{tabular}{|c|c|c|c|c|c|}

\hline
\multirow{3}{*}{Superv.} & \multirow{3}{*}{Method} & \multicolumn{4}{c|}{EndoVis} \\
\cline{3-6}
& & \multicolumn{2}{c|}{2017} & \multicolumn{2}{c|}{2018} \\
\cline{3-6}
& & AP@0.5 & AP@0.7 & AP@0.5 & AP@0.7\\
\hline

        \multirow{2}{*}{GT}
         & MRCNN & 76.11 & 61.87 & 75.01 & 63.12 \\                                    
         \cline{2-6} 
         & SAF-IS & \textbf{88.40} & \textbf{72.12} & \textbf{78.57} & \textbf{66.00}  \\ 
\hline
\hline

        \multirow{2}{*}{CC\textsubscript{M}}
         & MRCNN & 71.26 & 55.98 & 73.99 & 60.04  \\                                    
         \cline{2-6} 
         & SAF-IS & \textbf{85.36} & \textbf{63.70} & \textbf{75.92} & \textbf{61.08}  \\                                            
\hline
\hline
        \multirow{2}{*}{CC\textsubscript{F}}   
         & MRCNN & 63.81 & 44.99 & 62.48 & 42.31  \\                                    
         \cline{2-6} 
         & SAF-IS & \textbf{81.31} & \textbf{56.14} & \textbf{71.01} & \textbf{49.17}  \\                                        
\hline

\end{tabular}
\caption{Tool instantiation results for the proposed SAF-IS approach and Mask-RCNN on EndoVis 2017 and 2018 datasets, trained according to three modalities: fully-supervised (GT) and unsupervised using Connected Component labelling of manually annotated masks (CC\textsubscript{M}) and FUN-SIS predicted masks (CC\textsubscript{F}).}
\label{tab:SAF-IS:inst_results}
\end{table*}

\subsection{Design Choices \& Training Details}
\label{Sect:SAF-IS:exp:design}

\indent \textbf{Tool instantiation}: the instantiation model is implemented as a U-Net architecture with SegFormer encoder \citep{xie2021segformer},  available from the \textit{Segmentation Models} library in PyTorch. Training was carried out for 60 epochs using the Adam optimizer with a learning rate equal to 1e-3 and a batch size of 32, applying standard photometric and geometric augmentations from the \textit{Albumentation} library to the original images, resized to a $256\times256$ resolution. During inference, centroids were selected by overlapping the predicted displacement field with a square grid of $32\times32$ resolution (i.e. each grid square of $8\times8$ pixel dimension); a threshold $\varepsilon_C$ of 5 was used to select centroid squares (i.e. squares with a per-pixel average of at least 5 displacement vectors pointing at them were selected as centroids). The impact of grid resolution and threshold value is investigated in Section \ref{Sect:SAF-IS:abl}.\\
\indent \textbf{Feature representation learning}: the feature extractor network is implemented as a ResNet-50 architecture. Each instance mask is multiplied by the output of the $conv3\_4$ layer. Instance-wise features are obtained by applying a global average pooling to the output of the $conv5\_3$ layer, having 2048 feature channels. Training was carried out for 80 epochs using the Adam optimizer with a learning rate equal to 5e-5 and a batch size of 64, applying standard photometric and geometric augmentations to the original images, resized to a $512\times512$ resolution. For the contrastive loss $L_{SCL}$ a temperature factor equal to 0.1 was used.\\
\indent \textbf{Instance classification}: for the main experiments (Section \ref{Sect:SAF-IS:res:class}), K-Means++ clustering algorithm was applied with a total number of clusters $N_{km}$ equal to 8 (therefore 8 instances were required to be labelled by a human user). While in the real scenario such assignment would be performed by a human operator, as discussed in Section \ref{Sect:SAF-IS:abl}, it was here automatically performed by associating to each prototype instance the semantic label of the ground truth instance of the same frame having the maximum overlap according to the Intersection-over-Union metric.\\
The classification networks (Teacher, Student) were implemented as a 2-layer fully-connected network, with intermediate feature size of 512 and batch normalization. Training was carried out for 40 epochs using the Adam optimizer with a learning rate equal to 1e-4 and a batch size of 128, applying standard photometric and geometric augmentations to the original images, resized to a $512\times512$ resolution.

\section{Experiments and Results Analysis}
\label{Sect:SAF-IS:res}

We now present the experimental validation of the proposed SAF-IS framework, and compare it with state-of-the-art approaches. Tool instantiation results and complete instance segmentation results are separately presented in Sections \ref{Sect:SAF-IS:res:inst} \& \ref{Sect:SAF-IS:res:class}, respectively.

\LetLtxMacro{\originalcite}{\citep}
\def\tablecite#1#{%
  \def\pretablecite{#1}%
  \tableciteaux}
\def\tableciteaux#1{%
  \tiny \expandafter\originalcite\pretablecite{#1}%
}
\AtBeginEnvironment{table}{\let\citep\tablecite}

\begin{table*}[t]

\centering
\begin{tabular}{|c|c|c|c||c|c|c|c|c|}

\hline
\multirow{3}{*}{Method} & \multicolumn{6}{c|}{Supervision Type} & \multicolumn{2}{c|}{EndoVis} \\
\cline{2-9}
& \multicolumn{3}{c||}{Pixel-level} & \multicolumn{3}{c|}{Weak} & \multirow{2}{*}{2017} & \multirow{2}{*}{2018} \\
\cline{2-7} 
& \small S & I & B & P & FW & SW &  &  \\
\hline
Ternaus\citep{shvets2018automatic} & \checkmark & \phantom{0} & \phantom{0} & \phantom{0} & \phantom{0} & \phantom{0} & 35.27 & / \\ \hline
MF-TN\textsuperscript{\textdagger}\citep{jin2019incorporating} & \checkmark & \phantom{0} & \phantom{0} & \phantom{0} & \phantom{0} & \phantom{0} & 37.35 & / \\ \hline
DMF-TN\textsuperscript{\textdagger}\citep{zhao2020learning} & \checkmark\textsubscript{30\%} & \phantom{0} & \phantom{0} & \phantom{0} & \phantom{0} & \phantom{0} & \textbf{45.83} & / \\ \hline
DMF-TN\textsuperscript{\textdagger}\citep{zhao2020learning} & \checkmark\textsubscript{20\%} & \phantom{0} & \phantom{0} & \phantom{0} & \phantom{0} & \phantom{0} & 43.71 & / \\ \hline
DMF-TN\textsuperscript{\textdagger}\citep{zhao2020learning} & \checkmark\textsubscript{10\%} & \phantom{0} & \phantom{0} & \phantom{0} & \phantom{0} & \phantom{0} & 33.64 & / \\ \hline
\hline
M\&C\textsuperscript{\textdagger\textdaggerdbl}\citep{kurmann2021mask} & \checkmark & \checkmark & \phantom{0} & \phantom{0} & \phantom{0} & \phantom{0} & \textbf{65.70} & / \\ \hline
ISI-Net\textsuperscript{\textdagger}\citep{gonzalez2020isinet} & \checkmark & \checkmark & \phantom{0} & \phantom{0} & \phantom{0} & \phantom{0} & 55.62 & 73.03 \\ \hline
MRCNN\citep{kong2021accurate} & \checkmark & \checkmark & \phantom{0} & \phantom{0} & \phantom{0} & \phantom{0} & 42.28 & / \\ \hline
Tra-SeTr\textsuperscript{\textdagger}\citep{zhao2022trasetr} & \checkmark & \checkmark & \phantom{0} & \phantom{0} & \phantom{0} & \phantom{0} & 60.04 & \textbf{76.20} \\ \hline
\hline

SAF-IS & \phantom{0} & \phantom{0} & \checkmark & \checkmark\textsubscript{0.3\%} & \phantom{0} & \phantom{0} & 43.86 & 56.62 \\ \hline
SAF-IS & \phantom{0} & \phantom{0} & \checkmark & \checkmark\textsubscript{0.3\%} & \checkmark & \phantom{0} &\textbf{ 53.73} & 63.38 \\ \hline
SAF-IS & \phantom{0} & \phantom{0} & \checkmark & \checkmark\textsubscript{0.3\%} & \phantom{0} & \checkmark & 52.64 & \textbf{63.57} \\ 
\hline
\hline

SAF-IS & \phantom{0} & \phantom{0} & \phantom{0} & \checkmark\textsubscript{0.3\%} & \phantom{0} & \phantom{0} & 30.47 & 54.08 \\ \hline
SAF-IS & \phantom{0} & \phantom{0} \phantom{0} & & \checkmark\textsubscript{0.3\%} & \checkmark & \phantom{0} & \textbf{45.86} & \textbf{58.03} \\ \hline
SAF-IS & \phantom{0} & \phantom{0} & \phantom{0} & \checkmark\textsubscript{0.3\%} & \phantom{0} & \checkmark & 42.41 & 57.75 \\ \hline

\end{tabular}
\caption{Instance segmentation results for the proposed SAF-IS approach, state-of-the-art methods on EndoVis 2017 and 2018 datasets. Supervision signals used by each approach are reported: pixel-level semantic labels (S, percentage of labelled data reported for semi-supervised approaches), pixel-level instance labels (I), required by fully-supervised instance segmentation approaches, binary segmentation masks (B, for SAF-IS, if not checked FUN-SIS predicted masks are used), prototype labels (P, 8 labels in total in these experiments, $\sim$0.3\% of total training instances), frame-wise tool presence labels (FW) and sequence-wise tool presence labels (SW). \textsuperscript{\textdagger} methods using temporal information at inference time. \textsuperscript{\textdaggerdbl} methods using additional tool-part annotations for training.}
\label{tab:SAF-IS:class_results}
\end{table*}

\begin{table}[!ht]
\centering
    
    \begin{subtable}[h]{3.5in}
        \centering
        \begin{tabular}{|c|c|c|c|c|c|}
        
            \hline
            \multicolumn{2}{|c|}{Augm.} & \multicolumn{4}{c|}{EndoVis}\\ \hline 
            \multirow{2}{*}{OV} & \multirow{2}{*}{PS} & \multicolumn{2}{c|}  {2017} & \multicolumn{2}{c|}{2018} \\ \cline{3-6}
            & & AP@0.5 & AP@0.7 & AP@0.5 & AP@0.7 \\ \hline
            \phantom{0} & \phantom{0} & 74.85 & 56.585 & 71.56 & 58.08\\ \hline
            \checkmark & \phantom{0} & 77.74 & 54.82 & \textbf{77.58} & 62.00\\ \hline
            \phantom{0} & \checkmark & 81.91 & 59.82 & 70.54 & 57.98\\ \hline
            \checkmark & \checkmark & \textbf{85.35} & \textbf{63.70} & 75.92 & \textbf{62.08}\\ \hline
            
        \end{tabular}
        \caption{}
    \end{subtable}

\hspace*{0.1em}

    \begin{subtable}[h]{3.5in}
        \centering
        \begin{tabular}{|c|c|c|c|c|c|}
        
            \hline
            \multicolumn{2}{|c|}{Augm.} & \multicolumn{4}{c|}{EndoVis}\\ \hline 
            \multirow{2}{*}{OV} & \multirow{2}{*}{PS} & \multicolumn{2}{c|}  {2017} & \multicolumn{2}{c|}{2018} \\ \cline{3-6}
            & & AP@0.5 & AP@0.7 & AP@0.5 & AP@0.7 \\ \hline
            \phantom{0} & \phantom{0} & 67.82 & 47.86 & 65.23 & 43.94\\ \hline
            \checkmark & \phantom{0} & 71.80 & 45.69 & \textbf{71.91} & 48.99\\ \hline
            \phantom{0} & \checkmark & 72.41 & 49.42 & 67.99 & 47.12\\ \hline
            \checkmark & \checkmark & \textbf{81.31} & \textbf{56.14} & 71.01 & \textbf{49.16}\\ \hline
            
        \end{tabular}
        \caption{}
    \end{subtable}

\caption{Results of the ablation study on unsupervised instrument instantiation from manually annotated (a) and FUN-SIS predicted (b) binary masks, highlighting the separate and combined impact of: masking of potentially overlapping
instances (OV) and pasting of random tool instances (PS).}
\label{tab:SAF-IS:inst_abl}
\end{table}

\subsection{Tool Instantiation}
\label{Sect:SAF-IS:res:inst}

In order to analyze tool instantiation quality, we evaluate results according to a class-agnostic Average-Precision metric, computed for two values of threshold Intersection-Over-Union (IoU): AP@0.5 (50\%), AP@0.7 (70\%). We present results obtained by our unsupervised approach using, as binary masks, both manual annotations (SAF-IS CC\textsubscript{M}) and unsupervised FUN-SIS predictions (SAF-IS CC\textsubscript{F}). In addition, we report results for the instantiation model trained in a fully-supervised manner on the ground truth displacement field (SAF-IS GT). As, to the best of our knowledge, no other work has previously attempted unsupervised instantiation of binary tool masks, we compare our solution against a Mask-RCNN baseline, trained under the same fully-supervised (MRCNN GT) and unsupervised modalities (MRCNN CC\textsubscript{M}, \mbox{MRCNN CC\textsubscript{F}}). However, as Mask-RCNN is an anchor-based approach, the local masking for automatically identified overlapping tools ($M^{OV}$, described in Section \ref{Sect:SAF-IS:method:inst}), is not easily implementable, and would require substantial architectural modifications which are beyond the scope of this work. Therefore we limit the augmentation strategy for unsupervised Mask-RCNN experiments to instance pasting, described in Section \ref{Sect:SAF-IS:method:inst}. \\
 \indent Results presented in Table \ref{tab:SAF-IS:inst_results} show how our proposed solution outperforms Mask-RCNN across both datasets and for all three training modalities. A similar result for the fully-supervised training modality was already presented in \citep{kurmann2021mask}. These experiments highlight the benefits of tool instantiation based on direct centroid regression, beyond full-supervision, for the unsupervised setting. Indeed, the unsupervised SAF-IS solution using binary annotated masks (SAF-IS CC\textsubscript{M}) closely follows the fully-supervised one (SAF-IS GT), with an average gap of -$\Delta$3.3\% AP@0.5 across the two datasets. In addition, the greatest performance gap between SAF-IS and Mask-RCNN is found when using FUN-SIS binary masks to train (CC\textsubscript{F}): +$\Delta$17.5\% AP@0.5 and +$\Delta$11.15\% AP@0.7, in the EndoVis 2017 dataset. This result shows how our solution is particularly suitable to handle a noisy supervision signal. Finally, the performance gap between SAF-IS CC\textsubscript{F} and SAF-IS CC\textsubscript{M} is significantly smaller for the AP@0.5 metric (-$\Delta$4.48\% on average across the two datasets) compared to the  AP@0.7 metric (-$\Delta$9.74\%). This can be attributed to the lower quality of FUN-SIS binary segmentation masks, causing a performance drop when a high IoU threshold is used: the lower 50\% IoU threshold, instead, being less affected by possible inaccuracies in the binary segmentation masks, highlights the high instantiation quality.


\subsection{Tool Instance Segmentation}
\label{Sect:SAF-IS:res:class}
\begin{figure*}[!ht]
    \centerline{\includegraphics[width=\textwidth]{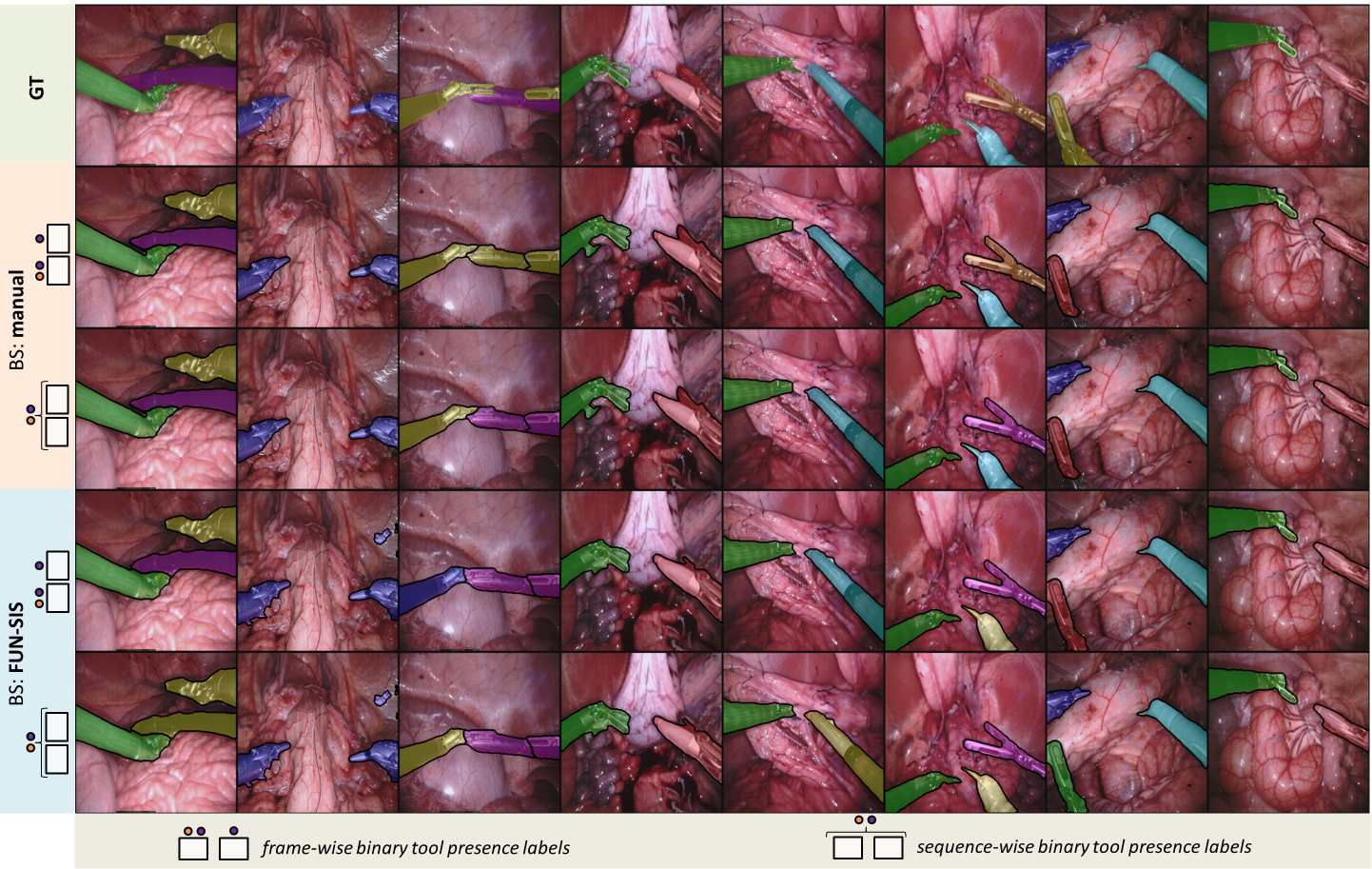}}
    \caption{Qualitative segmentation results from the EndoVis2017 dataset, highlighting, for our SAF-IS approach, the source of binary segmentation masks (BS) and the type of binary tool presence labels (frame-wise or sequence-wise). All the SAF-IS results are obtained using 8 prototype labels. Row 1: ground truth; rows 2-5: SAF-IS Student trained on (2) manually annotated binary masks and \textit{frame-wise} tool presence labels, (3) manually annotated binary masks and \textit{sequence-wise} tool presence labels, (4) FUN-SIS predicted binary masks and \textit{frame-wise} tool presence labels, (3) FUN-SIS predicted binary masks and \textit{sequence-wise} tool presence labels.}
    \label{fig:SAF-IS:qual_17}
\end{figure*}

\begin{figure*}[!ht]
    \centerline{\includegraphics[width=\textwidth]{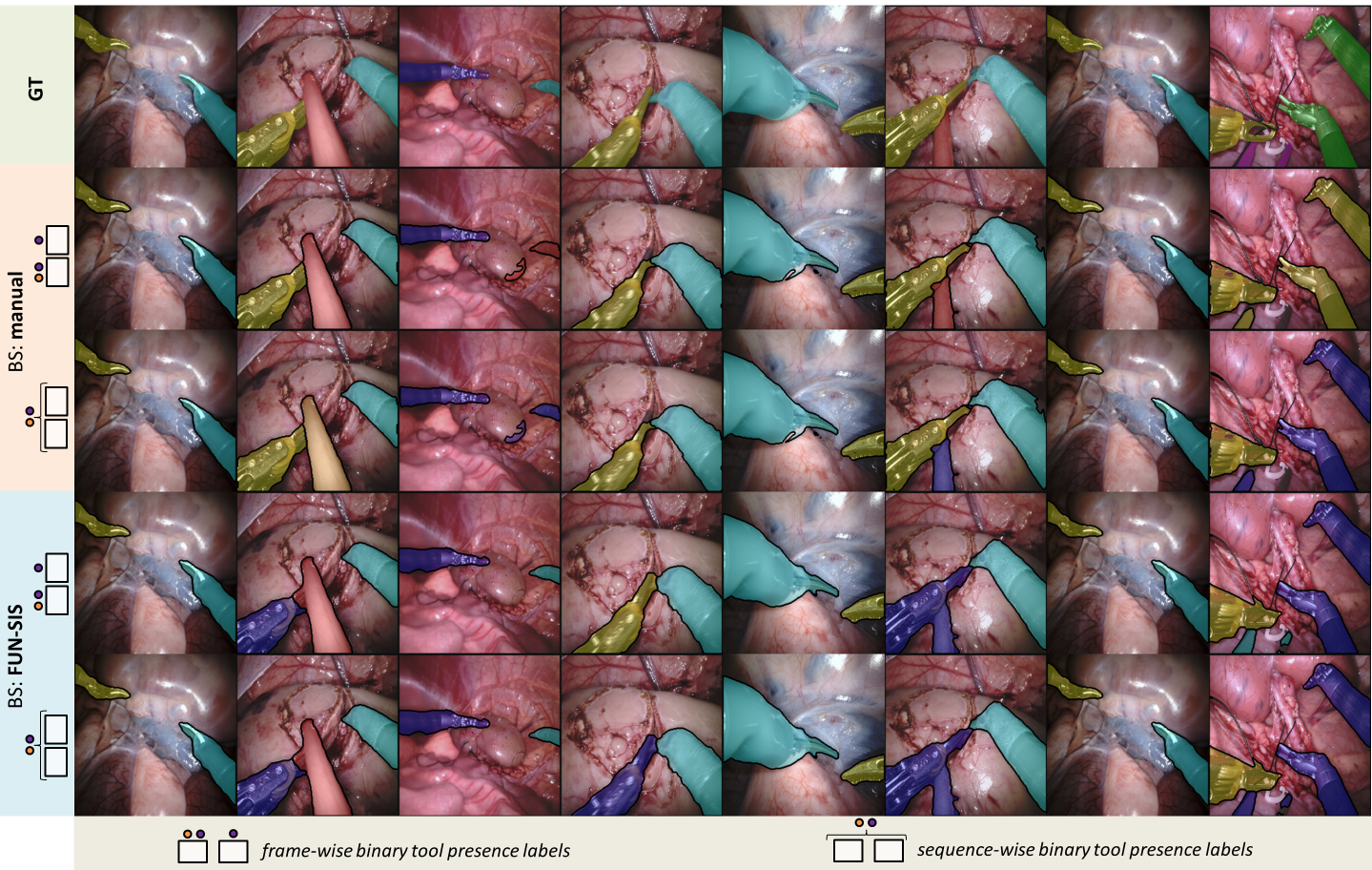}}
    \caption{Qualitative segmentation results from the EndoVis2018 dataset. Ground truth and SAF-IS Student results presented in the same order as Figure \ref{fig:SAF-IS:qual_17} above.}
    \label{fig:SAF-IS:qual_18}
\end{figure*}

In order to evaluate instance segmentation results, and compare them with other state-of-the-art segmentation approaches, we adopt the commonly used IoU EndoVis challenge metric defined in \citep{gonzalez2020isinet}. It is worth noticing that such metric treats the segmentation problem as pixel-wise classification, without providing information about instantiation quality. 
Table \ref{tab:SAF-IS:class_results} reports results of our SAF-IS framework and for several state-of-the-art solutions. For each method the table highlights the type of supervision used for training. State-of-the-art approaches are all trained in a fully-supervised manner using pixel-level semantic annotations (S), in combination with pixel-level instance annotations for instance segmentation methods (I). Our SAF-IS framework does not require pixel-level semantic or instance annotations to train, relying instead only on prototype instance labels (P) - 8 for the experiments reported in this Table - and weak labels, in the form of \textit{frame-wise} (FW) or \textit{sequence-wise} (SW) tool presence labels (results for both modalities are reported). In addition SAF-IS can be trained using manually annotated binary masks (B) if available, or rely on the predictions of the unsupervised FUN-SIS approach (results for both modalities are also reported).\\
\indent Results presented in Table \ref{tab:SAF-IS:class_results} show that our SAF-IS approach, trained using only binary tool presence labels and 8 prototype labels, outperforms fully-supervised and semi-supervised solutions adopting a semantic segmentation problem formulation (Ternaus, MF-TN, DMF-TN), despite not requiring any spatial annotation. On the EndoVis 2017 dataset our solution also outperforms a standard Mask-RCNN (MRCNN), trained on manually annotated segmentation masks and bounding-boxes for ground truth instantiation. In addition to pixel-level semantic and instance annotations, the solutions outperforming our SAF-IS approach also rely on temporal information during inference (\textdagger) and additional tool-part segmentation annotations (\textdaggerdbl). It is worth noticing that temporal modelling is a natural extension for SAF-IS, as tool tracking information is already extracted as part of the instance-wise feature learning step.  Finally, a comparison between SAF-IS models trained on \textit{frame-wise} (FW) and \textit{sequence-wise} (SW) binary tool presence labels, shows the effectiveness of our teacher-student solution to extract a reliable supervision signal from the automatically obtainable \textit{sequence-wise} labels, with an average gap between the two of less than 1.2\% IoU, across datasets and binary mask sources.
Qualitative results are shown in Figures \ref{fig:SAF-IS:qual_17} \& \ref{fig:SAF-IS:qual_18} at the end of the manuscript.

\begin{figure*}[t]
    \centerline{\includegraphics[width=5.3in]{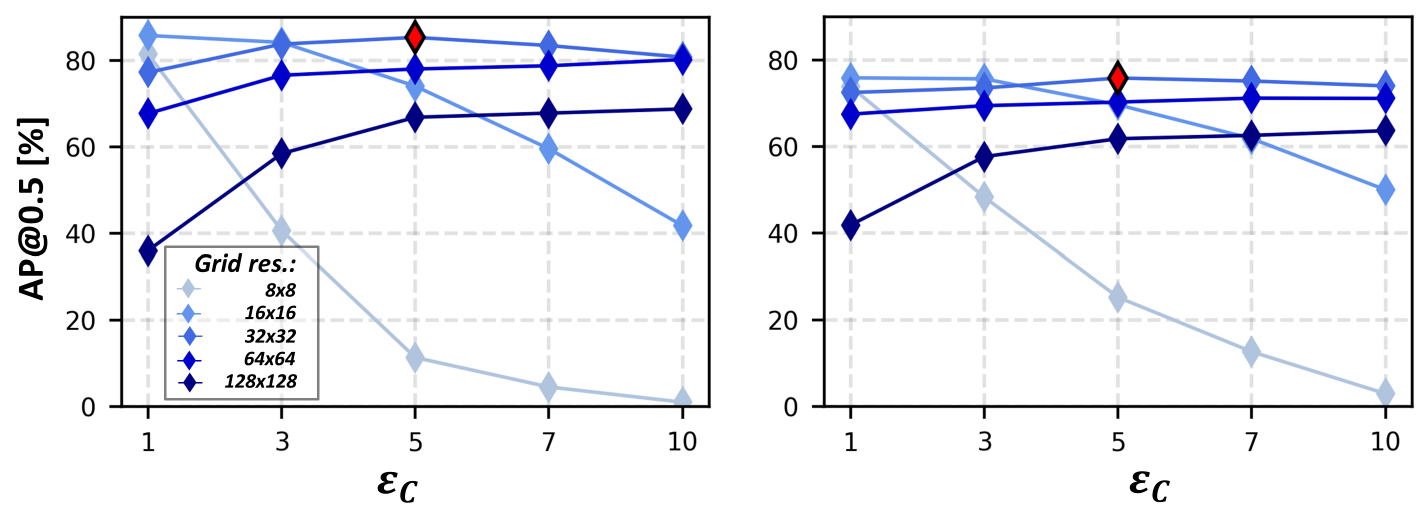}}
    \caption{Impact of grid square resolution and threshold value $\varepsilon_C$ on the tool instantiation quality for the EndoVis2017 dataset (left) and EndoVis2018 dataset (right). The combination used in our main experiments is highlighted in red.}
    \label{fig:SAF-IS:res_abl_quant}
\end{figure*}

\begin{figure*}[t]
    \centerline{\includegraphics[width=0.9\textwidth]{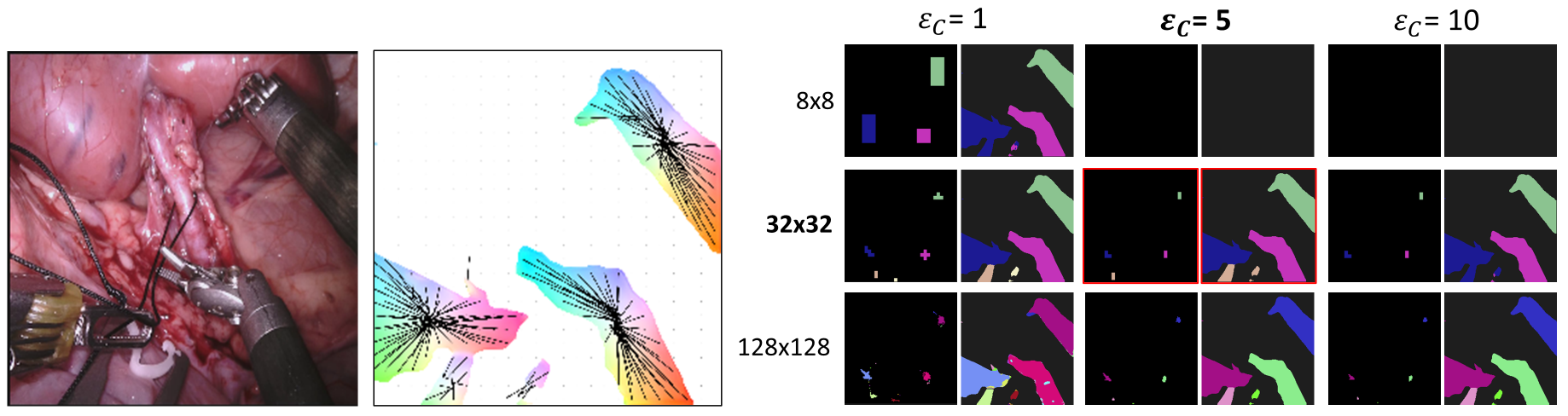}}
    \caption{From left to right, original image, predicted displacement field, and examples of centroid regions and instantiation masks for different combinations of grid resolution and threshold $\varepsilon_C$. Mask colors indicate the ID assigned to the tube each instance belongs to. The combination adopted in our main experiments is highlighted in red.}
    \label{fig:SAF-IS:res_abl_qual}
\end{figure*}

\section{Ablation Studies}
\label{Sect:SAF-IS:abl}

In order to provide a deeper insight into the SAF-IS framework, we now present and discuss ablation studies on three critical design choices: the augmentation strategy for tool instantiation, the inference parameters for tool instantiation and the number of prototype labels required for instance classification. \\

\subsection{Tool Instantiation Augmentation Strategy}
In order to train the displacement network for instrument instantiation, a pseudo-supervision signal is generated from the binary masks using a Connected Component algorithm. Such signal is subsequently refined by 1) preventing training on potentially overlapping instances (OV) and 2) pasting random tool instances (PS) to artificially simulate the case of overlapping instances (Section \ref{Sect:SAF-IS:method:inst}).\\
Table \ref{tab:SAF-IS:inst_abl} provides results of an ablation study exploring different combinations of the two augmentation strategies. Such results show the effectiveness of the two augmentation strategies, and of their simultaneous use. In the case of binary annotated masks, instance masking (OV) provides an average improvement of +$\Delta$4.46\% AP@0.5 and +$\Delta$1.08\% AP@0.7 across the two datasets, compared to the setting where no augmentation is used; instance pasting (PS) provide an average improvement of +$\Delta$3.03\% AP@0.5 and +$\Delta$1.56\% AP@0.7; the two strategies combined provide an average improvement of +$\Delta$7.02\% AP@0.5 and +$\Delta$5.55\% AP@0.7. On the EndoVis 2018 dataset, paste augmentation appears less effective: this could be due to the fact that several frames in it present at least 4 separate tool instances, making the additional pasting redundant, and potentially detrimental as frames can become too cluttered.\\

\begin{figure*}[!ht]
    \centerline{\includegraphics[width=0.9\textwidth]{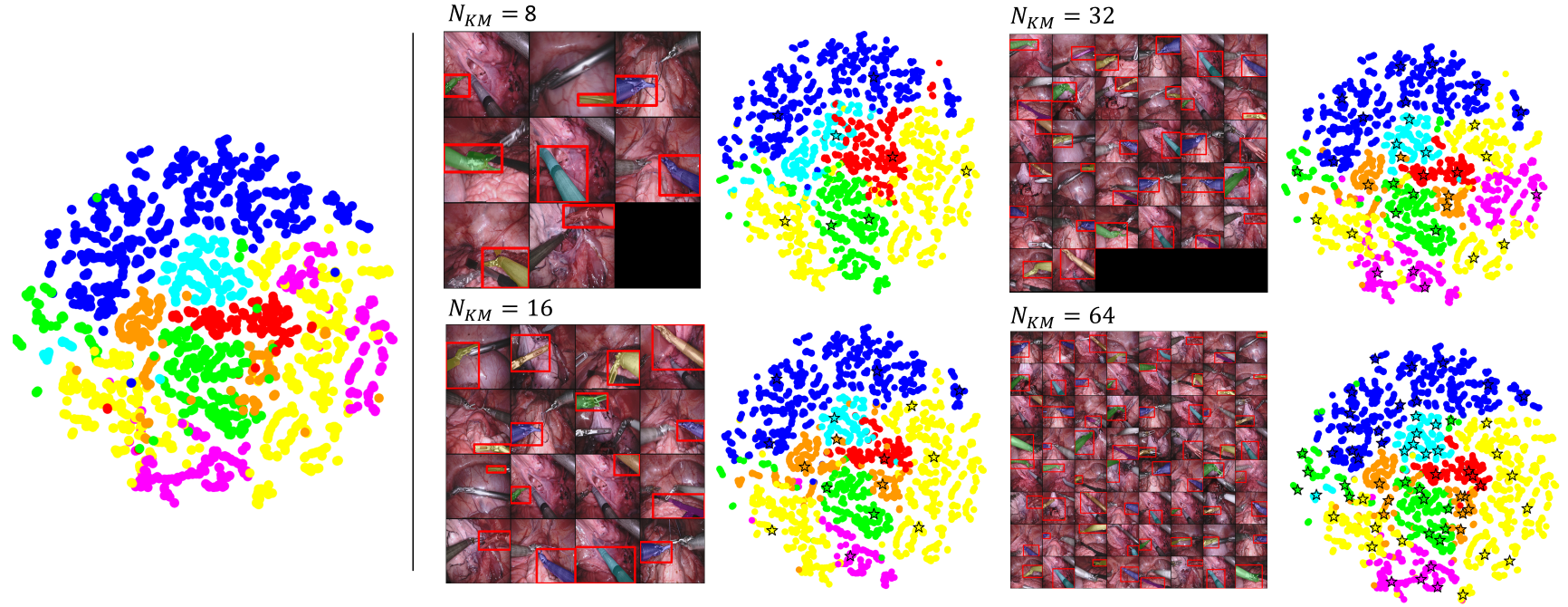}}
    \caption{Left: visualization of the learnt feature representations of the EndoVis 2017 training set instances, projected in the 2D space using t-SNE algorithm \citep{van2008visualizing}. Each instance point is colored according to the corresponding ground truth tool class. Right: K-Means++ clustering and prototype labels obtained using different number of clusters $N_{km}$; projected features and prototype instances are colored accordingly to the corresponding prototype labels.}
    \label{fig:SAF-IS:clusters}
\end{figure*}

\begin{table*}[!ht]
\centering

\begin{subtable}[ht]{0.49\textwidth}
\centering
\begin{tabular}{|c|c||c||c|c|}

\hline
\multirow{2}{*}{$N_{km}$} & \multirow{2}{*}{K-Means} & \multirow{2}{*}{Teacher} & \multicolumn{2}{c|}{Student}\\ \cline{4-5} 
& & & SW & FW \\ \hline
8 & 43.86 & 45.66 & 52.64 & \textbf{53.73} \\ \hline
16 & 38.52 & 41.34 & 50.02 & 52.64 \\ \hline
32 & 42.33 & 45.26 & 51.23 & 52.44 \\ \hline
64 & \textbf{44.76} & \textbf{48.38} & \textbf{52.84} & 53.37 \\ \hline

\end{tabular}
\caption{}
\label{tab:SAF-IS:proto_abl_e17_M}

\end{subtable}
\hfill
\begin{subtable}[ht]{0.49\textwidth}
\centering
\begin{tabular}{|c|c||c||c|c|}

\hline
\multirow{2}{*}{$N_{km}$} & \multirow{2}{*}{K-Means} & \multirow{2}{*}{Teacher} & \multicolumn{2}{c|}{Student}\\ \cline{4-5} 
& & & SW & FW \\ \hline
8 & 30.47 & 30.88 & 42.21 & 45.86 \\ \hline
16 & \textbf{37.86} & \textbf{41.81} & \textbf{46.95} & 47.33 \\ \hline
32 & 32.49 & 36.40 & 46.40 & \textbf{48.00} \\ \hline
64 & 36.10 & 41.02 & 46.91 & 47.96 \\ \hline

\end{tabular}
\caption{}

\label{tab:SAF-IS:proto_abl_e17_F}

\end{subtable}
\hfill
\begin{subtable}[ht]{0.49\textwidth}
\centering
\begin{tabular}{|c|c||c||c|c|}

\hline
\multirow{2}{*}{$N_{km}$} & \multirow{2}{*}{K-Means} & \multirow{2}{*}{Teacher} & \multicolumn{2}{c|}{Student}\\ \cline{4-5} 
& & & SW & FW \\ \hline
8 & \textbf{56.6}2 & 56.80 & \textbf{63.57} & 63.38 \\ \hline
16 & 56.03 & 57.25 & 60.63 & 61.96 \\ \hline
32 & 56.11 & \textbf{57.48} & 62.80 & \textbf{64.76} \\ \hline
64 & 53.80 & 57.22 & 62.24 & 63.88 \\ \hline

\end{tabular}
\caption{}
\label{tab:SAF-IS:proto_abl_e18_M}

\end{subtable}
\hfill
\begin{subtable}[ht]{0.49\textwidth}
\centering
\begin{tabular}{|c|c||c||c|c|}

\hline
\multirow{2}{*}{$N_{km}$} & \multirow{2}{*}{K-Means} & \multirow{2}{*}{Teacher} & \multicolumn{2}{c|}{Student}\\ \cline{4-5} 
& & & SW & FW \\ \hline
8 & 54.08 & 54.14 & 57.75 & 58.03 \\ \hline
16 & \textbf{56.53} & \textbf{57.04} & 57.40 & 58.53 \\ \hline
32 & 55.53 & 55.86 & \textbf{58.45} & 59.48 \\ \hline
64 & 55.52 & 56.02 & 57.92 & \textbf{59.85} \\ \hline

\end{tabular}
\caption{}

\end{subtable}
\caption{Results of the ablation study investigating the impact of the number of clusters $N_{km}$ on final segmentation results, using, for instance classification, direct K-Means inference, Teacher predictions and Student predictions, trained using \textit{sequence-wise} (SW) or \textit{frame-wise} (FW) binary tool presence labels. Results obtained using a): manually annotated binary masks on the EndoVis2017 dataset, b): FUN-SIS predicted binary masks on the EndoVis2017 dataset, c): manually annotated binary masks on the EndoVis2018 dataset, d): FUN-SIS predicted binary masks on the EndoVis2018 dataset. Segmentation results were evaluated using the challenge IoU metric. The best results across the number of clusters are highlighted in bold.}
\label{tab:SAF-IS:proto_abl}
\end{table*}

\subsection{Tool Instantiation Inference Parameters}
In order to obtain instance masks, a square grid is overlapped to the predicted displacement field; centroid squares are then selected as the ones whose per-pixel average of vectors pointing inside them is greater than the threshold value $\varepsilon_C$. The grid resolution (equal to $32\times32$ in our main experiments) and the threshold $\varepsilon_C$ (equal to 5 in our main experiments) regulate the trade-off between precision and recall of the obtained instance masks. We experimentally evaluate the impact of the two parameters by varying them in a grid-like manner, with grid resolution in $[8,16,32,64,128]$ and $\varepsilon_C$ in $[1,3,5,7,10]$. Their different combinations are used to obtain instance masks from the same displacement fields. The AP@0.5 between the obtained masks and the ground truth instances is reported in Figure \ref{fig:SAF-IS:res_abl_quant} for both the EndoVis2017 and EndoVis2018 datasets.\\
\noindent The presented results, together with the qualitative results shown in Figure \ref{fig:SAF-IS:res_abl_qual}, clearly highlight the impact of the two parameters. For intermediate grid resolution values ($32\times32$, $64\times64$), the impact of $\varepsilon_C$ is minimal. However, as the grid solution decreases ($16\times16$, $8\times8$), an high value of $\varepsilon_C$ negatively affects the quality as instantiation, as the average convergence rate on large squares tends to be lower. This can be also observed from the qualitative instantiation results shown in Figure \ref{fig:SAF-IS:res_abl_qual}, top-right, where no candidate squares reach the threshold. Vice-versa, high grid resolution values ($128\times128$) tend to be more negatively affected by a low $\varepsilon_C$, as it leads to the identification of many false positive centroids (instantiation results from Figure \ref{fig:SAF-IS:res_abl_qual}, bottom-left).

\subsection{Prototype Labels Number}
In SAF-IS, the Teacher network is required to gather knowledge from the prototype labels, in order to be able to identify the correct ordered sets of weak labels used for Student training. Prototype labels, therefore can have a crucial influence on the quality of instance classification. In addition, they represent the manual annotation necessarily required by SAF-IS for training, as both binary tool masks and binary tool presence labels can be automatically obtained. Therefore we now present, in Table \ref{tab:SAF-IS:proto_abl}, the impact on the segmentation performance, of the number of clusters $N_{km}$ used for K-Means clustering, equal to the number of prototype labels assigned by a human operator. In order to provide a complete overview, we present segmentation results obtained via instance classification by direct K-Means inference, Teacher classifier prediction and Student classifier prediction, when trained using \textit{sequence-wise} or \textit{frame-wise} binary tool presence labels. In addition, Figure \ref{fig:SAF-IS:clusters} provides a visualization of the learnt feature distribution, the clustering process and the automatically selected prototype instances.\\
Result analysis provides different insights into the method. First of all, although a marginal improvement exists, increasing the number of prototype instances does not provide substantial performance gains for the Student network. This result may indicate that effective feature learning is a crucial methodological bottleneck, which cannot be solved by simply increasing the number of human-assigned labels. Secondly, the presented results highlight the consistent improvement in performance provided by the Student network, trained on the weak labels matched through the Teacher model. Although the Teacher learns to substantially replicate K-Means clustering classification, as shown by their similar performance, this is enough to perform a good weak label matching, responsible for Student's superior performance.

\section{Discussion}
The results presented in Sections \ref{Sect:SAF-IS:res} \& \ref{Sect:SAF-IS:abl} confirm the soundness of the proposed SAF-IS framework for instance segmentation. Our solution trains on endoscopic videos paired with binary segmentation masks, potentially obtained in an unsupervised way, and is designed to incorporate binary tool presence labels, either \textit{frame-wise} or \textit{sequence-wise}. Human annotation effort can here be limited to labelling a tiny set of prototype instances, automatically selected by our approach, with inexpensive classification labels: the ablation study presented in Section \ref{Sect:SAF-IS:abl} shows that the size of such set can be reduced to 8 instances ($\sim$0.26\% of the total number of training instances), with no significant performance drop. This result goes significantly beyond existing semi-supervised solutions like \citep{zhao2020learning}, where a significant set of frames (up to 30\%) needs to be labelled with pixel-level annotations, while still providing inferior segmentation performance. Indeed, our complete spatial annotation-free solution, using FUN-SIS predicted binary masks for training, outperforms fully-supervised and semi-supervised semantic segmentation approaches like MF-TN and DMF-TN by a consistent margin on the EndoVis 2017 dataset. Furthermore, our SAF-IS framework effectively incorporates \textit{sequence-wise} binary tool presence labels, commonly overlooked in the literature.
This small gap in performance between \textit{frame-wise} and \textit{sequence-wise} training modalities (Table \ref{tab:SAF-IS:class_results}), shows that \textit{sequence-wise} labels can be an effective source of supervision, while being completely free to collect.\\
Although a performance gap still exists with top-performing fully-supervised instance segmentation approaches, we believe there exist several directions of improvement to close such a gap. First of all, temporal modelling could be easily learnt from the already available tracking information, currently exploited only at training time for feature learning. Secondly, as highlighted by the ablation study on cluster number, feature learning represents a crucial methodological bottleneck: if the learnt feature representations are sub-optimal, the unsupervised clustering may fail to separate tools belonging to different classes, hindering the following classifier training. In the current implementation, feature learning is performed in a completely unsupervised way, with no help from external information. Weak information about binary tool presence may be included at this stage to perform a more informed positive and negative feature sampling.\\
\begin{figure*}[ht]
    \centerline{\includegraphics[width=\textwidth]{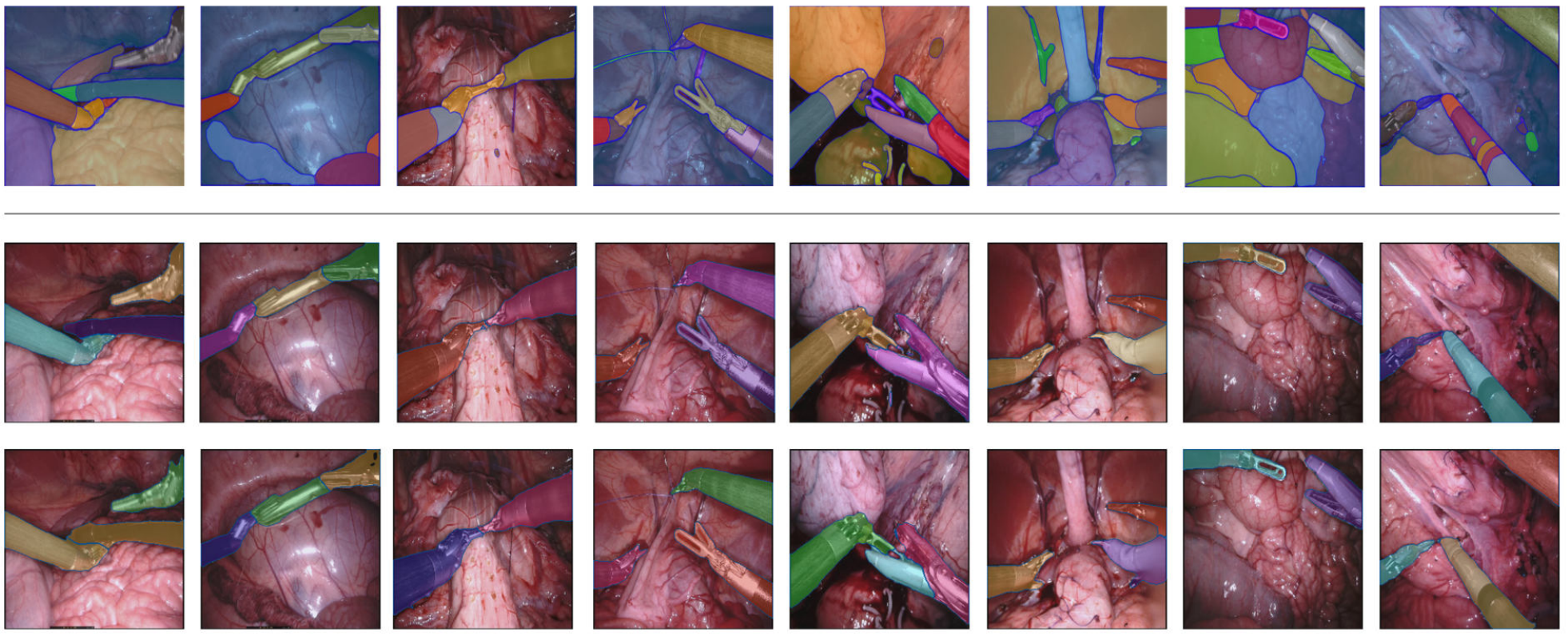}}
    \caption{Top row: SAM \citep{kirillov2023segment} segmentation results on the EndoVis2017 dataset. Central row: SAF-IS instantiation results obtained from binary manually annotated masks. Bottom row: SAF-IS instantiation results obtained from FUN-SIS predicted binary masks. Each instance is colored using a random color, which is not meant to represent tool type classes.}
    \label{fig:SAF-IS:sam}
\end{figure*}
\indent In addition to these direct improvements, SAF-IS, not requiring pixel-level labels, can leverage recent break-through solutions like SAM (Segment Anything Model, \cite{kirillov2023segment}) to directly obtain instance-wise masks for the following feature learning and tool classification training. Figure \ref{fig:SAF-IS:sam} shows qualitative results from SAM (without text prompts, not yet released at the time of this submission) on the EndoVis2017 dataset, compared to SAF-IS predictions. Even if SAM segmentation results are currently over-segmenting tools, breaking them up into individual parts, our SAF-IS instantiation predictions could be used to group these parts, exploiting the high-quality boundary segmentation that SAM can already provide.\\
In conclusion, SAF-IS major contribution lies in its ability to lift the need for spatial annotation of the training data. This may open up new research directions aimed at better exploiting human annotation effort, for example by focusing it on particularly representative or challenging samples.

\section{Conclusion}
In this work we developed and validated SAF-IS, a Spatial Annotation Free framework for Instance Segmentation of surgical instruments. The proposed framework embraces recent breakthrough solutions for unsupervised binary segmentation, building on top of them to perform instance segmentation without requiring pixel-level semantic or instance annotations to train. Instead, SAF-IS exploits the binary tool masks to learn to encode each instance in a compact feature representation, and solves the instance classification problem by relying on cheaply obtainable binary tool presence labels. A supplementary video highlighting crucial methodological aspects and providing additional qualitative results is available at \url{https://vimeo.com/860204311}. In conclusion, we hope this work can show the potential of prior knowledge and weakly-supervised training for tool instance segmentation, encouraging the search for alternatives to full-supervision for increasingly complex surgical computer vision tasks.

\section*{Acknowledgments}
This work was supported by the ATLAS project. The ATLAS project has received funding from the European Union’s Horizon 2020 research and innovation programme under the Marie Sklodowska-Curie grant agreement No. 813782. This work was also partially supported by French State Funds managed by the Agence Nationale de la Recherche (ANR) through the Investissements d’Avenir Program under Grant ANR-11-LABX-0004 (Labex CAMI) and Grant ANR-10-IAHU-02 (IHU-Strasbourg), by French state funds managed by the ANR under references ANR-20-CHIA-0029-01 (National AI Chair AI4ORSafety) and ANR-18-CE19-0012, and by BPI France under reference DOS0180017/00 (project 5G-OR).

\bibliographystyle{model2-names.bst}\biboptions{authoryear}
\bibliography{references}

\end{document}